\documentclass[preprint,3p,twocolumn]{elsarticle}

\usepackage{gensymb}
\usepackage{amssymb}
\usepackage{amsmath}
\usepackage[bookmarks=true,colorlinks,urlcolor=blue]{hyperref}
\usepackage[capitalise]{cleveref}
\usepackage{mathtools}
\usepackage{makecell}

\usepackage[dvipsnames]{xcolor}
\usepackage{graphicx}
\usepackage[strings]{underscore}

\usepackage{blindtext}
\usepackage{subcaption}

\usepackage[numbers]{natbib}

\newcommand{\bmat}[1]{\begin{bmatrix} 
#1
\end{bmatrix}}

\DeclareMathOperator*{\argmin}{arg\,min}

\definecolor{edithighlight}{RGB}{0, 220, 20}
\newcommand{\journaledit}[1]{#1}

\newcommand{\until}[1]{\{1,\dots, #1\}}

\newcommand{\norm}[1]{\left\lVert#1\right\rVert}

\newcommand{\reals}{\mathbb{R}}

\newcommand{\ballo}[2]{B\left( {#1}, {#2} \right)}

\DeclareMathOperator{\sgn}{sgn}

\DeclareMathOperator{\sgnz}{sgnz}

\newcommand{\timedelay}{t_d}

\newcommand{\rb}{\mathbf{r}}

\newcommand{\Nset}{\mathcal{N}}

\newcommand{\pos}[1]{\rb_{#1}}

\newcommand{\vel}[1]{\dot{\rb}_{#1}}

\newcommand{\rpos}[1]{\Delta \pos{#1}}

\newcommand{\rvel}[1]{\Delta \vel{#1}}

\newcommand{\acc}[1]{\ddot{\rb}_{#1}}

\newcommand{\amax}{a_{\text{max}}}

\newcommand{\vmax}{v_{\text{max}}}

\newcommand{\lr}{\ell_r}

\newcommand{\cscale}{c_r}

\newcommand{\Fatness}{\Phi}
\newcommand{\Tangentness}{\tau}

\newcommand{\Fatmean}{\overline{\Fatness}}
\newcommand{\Tangmean}{\overline{\Tangentness}}

\newcommand{\convqual}{\lambda}

\newcommand{\udesire}[1]{\mathbf{u}_{#1}^*}

\newcommand{\uactual}[1]{\mathbf{u}_{#1}}

\newcommand{\rmin}{r_{\text{min}}}
\newcommand{\rmax}{r_{\text{max}}}

\newcommand{\possafe}{\mathbf{r}_{\text{safe}}}

\DeclareMathOperator*{\clip}{\mathbf{clip}}

\newcommand{\fnocrash}{\mathbf{F}_{\cscale}}

\newcommand{\fnocrashargs}{\fnocrash(\udesire{i}, \pos{i}, \vel{i}, \lbrace (\pos{j}, \vel{j}) \rbrace_{j \in \Nset_i})}

\newcommand{\ORCA}[2]{ORCA_{#1|#2}^{\cscale}}

\newcommand{\orcame}[2]{ORCA_{#1}^{#2}}

\newcommand{\vsafe}{\mathbf{v}_{\text{safe}}}

\begin{document}

\title{The impact of catastrophic collisions and collision avoidance on a swarming behavior}

\author[1]{Chris Taylor}
\ead{ctaylo3@gmu.edu}
\author[1]{Cameron Nowzari}
\ead{cnowzari@gmu.edu}

\address[1]{George Mason University, 4400 University Dr, Fairfax, VA 22030}

\date{\today}

\begin{abstract}

Swarms of autonomous agents are useful in many applications due to their ability to accomplish tasks in a decentralized manner, making them more robust to failures. 
Due to the difficulty in running experiments with large numbers of hardware agents, researchers often make simplifying assumptions and remove constraints that might be present in a real swarm deployment. 
While simplifying away some constraints is tolerable, we feel that two in particular have been overlooked: one, that agents in a swarm take up physical space, and two, that agents might be damaged in collisions. 
Many existing works assume agents have negligible size or pass through each other with no added penalty. 
It seems possible to ignore these constraints using collision avoidance, but we show using an illustrative example that this is easier said than done. 
In particular, we show that collision avoidance can interfere with the intended swarming behavior and significant parameter tuning is necessary to ensure the behavior emerges as best as possible while collisions are avoided. 
We compare four different collision avoidance algorithms, two of which we consider to be the best decentralized collision avoidance algorithms available. 
Despite putting significant effort into tuning each algorithm to perform at its best, we believe our results show that further research is necessary to develop swarming behaviors that can achieve their goal while avoiding collisions with agents of non-negligible volume.

\end{abstract}
 
\maketitle

\section{Introduction}

Swarms of biological agents are capable of impressive feats. 
Birds can flock in formations together to improve efficiency during migration or provide safety from predators, ants and termites can cooperatively build large complex structures, and bees and other social insects can cover wide areas foraging for resources \cite{bonabeau1999swarm}. 
More impressive is that all of this can be accomplished in a fully decentralized fashion: each animal only need interact with other animals locally or have no awareness of the global objective. 
The loss of any individual does not impair the overall swarm's objective in any way; in a sense the swarm is capable of ``self-healing'', making it extremely robust. 
Owing to the success of natural swarms, robotics researchers have sought to bring some of nature's capabilities to engineered platforms. 
\par 
Yet, there are large differences between natural and artificial swarms. 
Prototyping artificial swarms is very difficult and costly, so researchers usually make simplifying assumptions and just focus on one aspect of a swarm \journaledit{at a time}, and often in simulation. 
For instance, researchers might focus solely on the dynamic behavior such as flocking \cite{Vicsek1995} or rotational ``milling'' \cite{Carrillo2009}.
Other works consider specific constraints placed upon the swarm that might hinder its performance, for instance time-delayed communication \cite{Mier-Y-Teran-Romero2012}, limited communication graphs \cite{Szwaykowska2016}, collision avoidance \cite{DongEuiChang2003,VanDenBerg2011,Borrmann2015}, noise in actuation \cite{Lindley2013}, or limited sensing models \cite{Soria2019}. 
Swarms in nature, by contrast, are ``tuned'' through evolution with regards to all of these constraints \textit{simultaneously}. 
It is possible that we miss unintended side effects unless we explore the combined effects of these constraints. 
In particular, we believe there are two constraints whose combined effects with a swarming behavior have not been investigated thoroughly. 
\par 
\textbf{1. Agents use physical space. }
Many works assume agents have negligible size \cite{Szwaykowska2016,Zhang2015}, for instance \cite{Zhang2015} defines a collision as two agents occupying the \textit{exact} same position, \cite{Carrillo2009,Mier-Y-Teran-Romero2012,Szwaykowska2016} assume agents can pass right through one other, and many other works make no mention of the size of agents \cite{Ghapani2016,Zhan2013,Cao2012,Erdmann2005,DOrsogna2006,Szwaykowska2016}. 
However, as shown in an illustrative example in \cite{hamann2020guerrilla}, incorporating a physical size for agents completely changes how the performance scales with the number of agents, where adding too many incurs a performance drop as agents start to interfere with each other. 
\journaledit{Thus, some works which incorporate a physical size for agents go to great lengths to ensure the swarm still functions.} 
SDASH, for instance, is an algorithm that lets swarms form shapes while dynamically adjusting for the sizes and number of agents in the swarm \cite{Rubenstein2010}. 
A followup work focuses on optimizing the task assignment, in particular getting agents to choose goal waypoints physically close to them so they don't have to navigate through crowds of other agents \cite{Wang2020}. 
\par 
\textbf{2. Collisions can harm agents.}
Although it sounds obvious, this constraint is unnecessary when we consider ``soft'' agents like sheep \cite{Zuriguel2016} or fish \cite{Viscido2005}. 
Some works with robotic agents intentionally exploit collisions as a means of gathering information \cite{Mayya2017,Mulgaonkar2018}. 
However, in some hardware platforms, for instance the quadrotors used in \cite{chung,Vasarhelyi2018a}, collisions are unacceptable and would severely damage the agents involved. 
\par 
In some cases, combining the previous two constraints with a swarming behavior will have negligible effects, for instance if agents need to spread out to cover a large area then collisions are mostly irrelevant \cite{Arul2019,Beard2004}. 
But with other cases, such as the highly dynamic \textit{double-milling} behavior seen in \cite{Szwaykowska2016}, agents frequently encounter each other in opposing direction at high speeds and collisions would quickly destroy the swarm. 
\par 
The most obvious course of action to rectify this problem would be to add collision avoidance, but as we show in previous works \cite{taylor2020acc,taylor2020ants}, this is easier said than done and there is a delicate balancing act in parameter tuning needed to ensure the swarm still functions while minimizing collisions. 
Our latest work \cite{taylor2020ants} explores more sophisticated \textit{minimally invasive} collision avoidance techniques compared to our first work \cite{taylor2020acc}, but it is still a surface-level exploration. 
In particular, each collision avoidance algorithm has its own parameter to control how ``reckless'' or ``cautious'' each performs, but the parameter has different meanings and different units in every algorithm. 
It is possible that by not properly ``tuning'' each algorithm to perform at its best we unfairly represent each one. 
\par 
In this work, we continue the investigation of \cite{taylor2020acc,taylor2020ants} but we make significant efforts to tune each collision avoidance algorithm to the best of its ability by varying the ``cautiousness'' parameter described previously. 
We replicate the swarming algorithm from \cite{Szwaykowska2016} and then measure the interference caused by four different collision avoidance algorithms: the original potential-fields approach included with \cite{Szwaykowska2016}, a ``gyroscopic'' steering control \cite{DongEuiChang2003},  control barrier certificates \cite{Borrmann2015}, and Optimal Reciprocal Collision Avoidance (ORCA) \cite{VanDenBerg2011}. 
Despite our best efforts to tune each algorithm, there is still unavoidable interference from adding constraints of physical size and catastrophic collisions. 
In addition, the tuning process itself is problematic since the relationship between collision avoidance parameters and model performance appears difficult to predict. 
Our work shows that more research is needed to understand swarming behaviors under collision constraints.

\section{Problem Formulation}

We wish to understand the effects of imposing two constraints on the swarming algorithm from Szwaykowska et al.~\cite{Szwaykowska2016}: one, that agents have non-negligible physical size, and two, that agents are destroyed upon colliding. 
Naturally, we add a collision avoidance algorithm to allow the behavior to function while satisfying these constraints. 

\subsection{Individual Agent Model} \label{sec:probformulation:model}

To isolate just the effects of our constraints, we consider a simple agent model. 
Letting~$\pos{i} \in \reals^2$ be the position of agent~$i \in \until{N}$ in a swarm of $N$ agents, we consider the dynamics
\begin{align}\label{eqn:model}
    \acc{i}(t) = \uactual{i}(t)
\end{align}
with the following two constraints at all times~$t \in \reals_{\geq 0}$:
\\

\noindent \textbf{C1. Collisions destroy agents.} To be as general as possible, we do not assume that our collision-avoidance strategy is able to guarantee safety. Letting $r > 0$ represent the physical radius of the agents, to allow for the possibility of collisions while keeping the number of agents $N$ fixed we let collided agents ``respawn'' in a safe location with zero velocity, i.e. 
\begin{equation}
\begin{split}
    \norm{\pos{i}(t^-) - \pos{j}(t^-)} < 2r \implies \\ 
        \pos{i}(t^+) &= \possafe - 4 \bmat{r & r}^T \\ 
        \pos{j}(t^+) &= \possafe - 8 \bmat{r & r}^T \\ 
        \vel{i}(t^+) &= \vel{j}(t^+) = \mathbf{0}. 
\end{split}
\end{equation}
where $\possafe$ is the lower-left corner of the bounding box enclosing all the agents, or
\[
    \possafe = \bmat{ \min \lbrace x_i \rbrace & \min \lbrace y_i \rbrace }. 
\]
\cref{fig:snapshots:collisions} shows an example. 
Without this added effect, agents could simply destroy each other until there are trivially few agents, making the deployment too easy. 
In addition, we want to study the steady-state behavior of the swarm for a fixed number of agents $N$ and letting $N$ change throughout the course of the swarm's operation would add confusion to the results. 
We feel that simply counting collisions (e.g. as \cite{Viscido2005} does) without considering their disruption on the swarm's state is not as illustrative of a real swarm deployment. 
\\
\\
\noindent \textbf{C2. Limited acceleration.} Since some of the collision avoidance strategies we use are unconstrained in how much control effort they can apply to add safety, we require $\norm{u_i(t)} \leq \amax$ so that agents cannot ``cheat'' by applying infinite acceleration. 

\subsection{Desired Global Behavior: Ring State}
Given the model and constraints, we now introduce the controller we replicate from \cite{Szwaykowska2016} which achieves the ``ring state'' when the parameters are chosen correctly. 
Without collision avoidance, it is
\begin{equation}
\begin{split}
     \udesire{i} = &  \beta (v_0^2 - \norm{\vel{i}}^2) \vel{i} \\
     & + \frac{\alpha}{N-1} \sum_{j \in \until{N} \backslash \lbrace i \rbrace} \left(\pos{j}(t - \timedelay) - \pos{i}(t)\right)  .
\end{split}
\label{eqn:udesire}
\end{equation}

The input $\udesire{i}$ consists of two terms (in order):
keep the agent's speed at approximately $v_0 > 0$ with gain $\beta > 0$
and attract toward the position of other agents $\timedelay$ seconds in the past ($d$ for \emph{delay}), where $\alpha$ adjusts the attraction strength. 
If the input for each agent consists solely of \cref{eqn:udesire}, the formation looks like \cref{fig:snapshots:perfect}, where agents form possibly counter-rotating rings and pass through each other. 
As this is not realistic when agents have volume and can collide, we must add collision avoidance. 
\par 
We assume each agent has access to the local non-delayed states of its neighbors within some circular disc of radius $\lr$. 
We define this set as 
\begin{align}
\begin{split}
	\Nset_i = \lbrace  
	     j \in \until{N} \rbrace \backslash \lbrace i \rbrace \; | \; 
		 \norm{\pos{i} - \pos{j}} \leq \lr 
		\rbrace.
\end{split}
\label{eqn:nsetdef}
\end{align}
We then consider a collision avoidance ``wrapper'' function $\fnocrash$ which takes an agent's possibly unsafe input $\udesire{i}$, its state, the set of states of its local neighbors, then converts it to a safer input $\uactual{i}$, i.e. 
\[
	\fnocrashargs \rightarrow \uactual{i}. 
\]
The wrapper function $\fnocrash$ is parameterized by a variable $\cscale$ which controls the ``cautiousness'' of the collision avoidance, where higher values of $\cscale$ map to stronger corrections in the desired input $\udesire{i}$. 
We want to make sure whichever collision avoidance method we use for $\fnocrash$ obeys constraint C2 and cannot circumvent inertial constraints with infinite acceleration, so we define
\begin{align*}
	\clip(\mathbf{x}, a) = 
	\begin{cases}
		 \mathbf{x} & \norm{\mathbf{x}} < a, \\
		a \frac{\mathbf{x}}{\norm{\mathbf{x}}}	 & \text{otherwise.}
	\end{cases}. 
\end{align*}
which clips the magnitude of $\mathbf{x}$ to $a$. 

This results in the final dynamics equations of
\begin{align}
	\acc{i} &= \uactual{i} \nonumber \\
	\uactual{i} &= \clip \lbrack \fnocrashargs, \amax \rbrack. 
\label{eqn:uactual}
\end{align}

Note that the \textit{all-to-all} sensing in \cref{eqn:udesire} is delayed by $\timedelay$ seconds (e.g. as communicated over a network), whereas the \textit{local} sensing used for collision avoidance in \cref{eqn:uactual} incurs no delay (e.g. agents sense others without communication). 
Our interest is primarily in the effect of collisions and collision avoidance on the ring behavior \cite{Szwaykowska2016}, so we consider simple communication assumptions that allow the ring to perform at its best. 
This helps us study the degradation in quality solely under collision constraints, as opposed to combined effects of collisions and realistic communication. 
We suspect adding realistic communication would degrade the quality of the behavior even further.

\section{Methodology}

\begin{figure*}
	\centering
	\begin{subfigure}[t]{0.16\textwidth}
		\includegraphics[width=\linewidth]{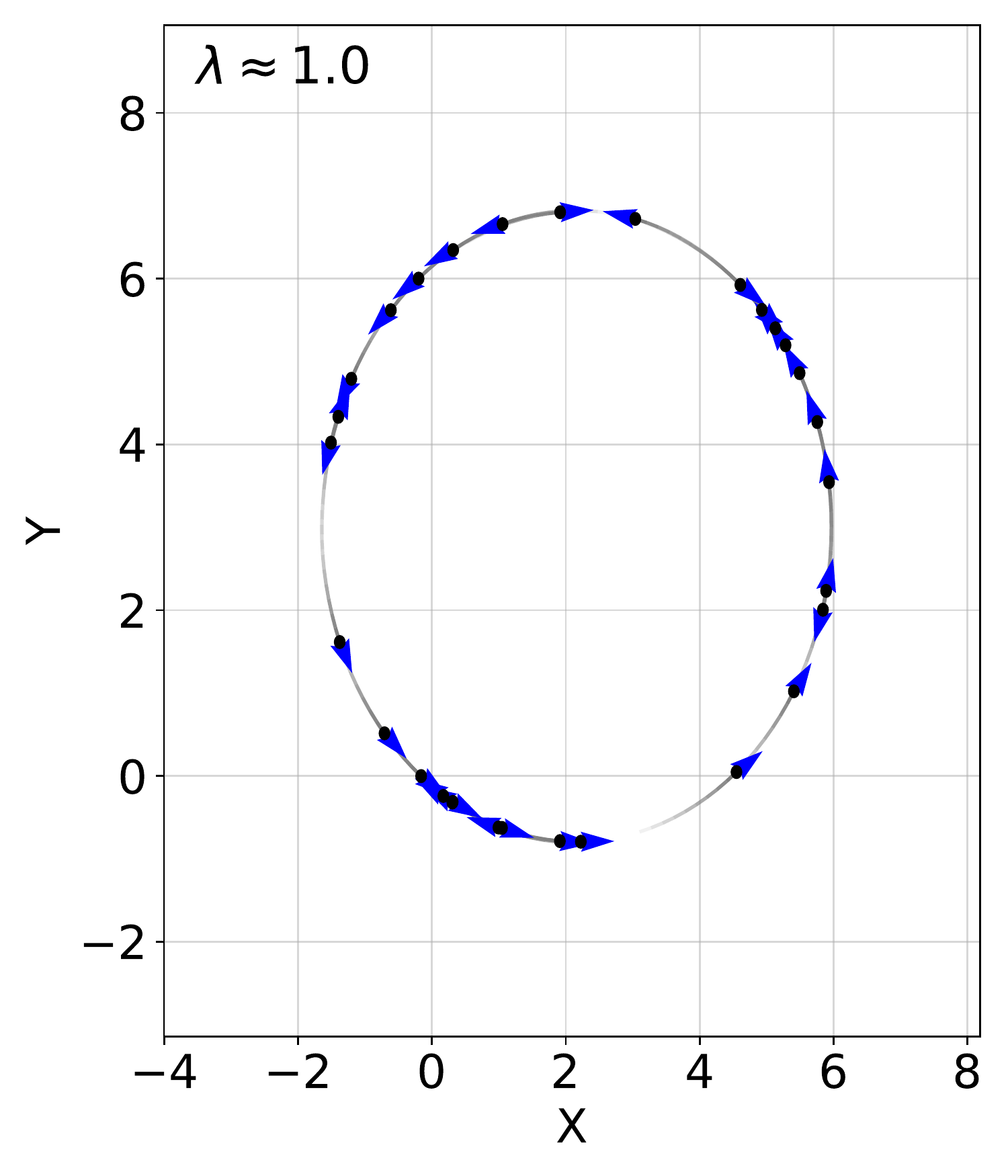}
	\end{subfigure}
	\begin{subfigure}[t]{0.16\textwidth}
		\includegraphics[width=\linewidth]{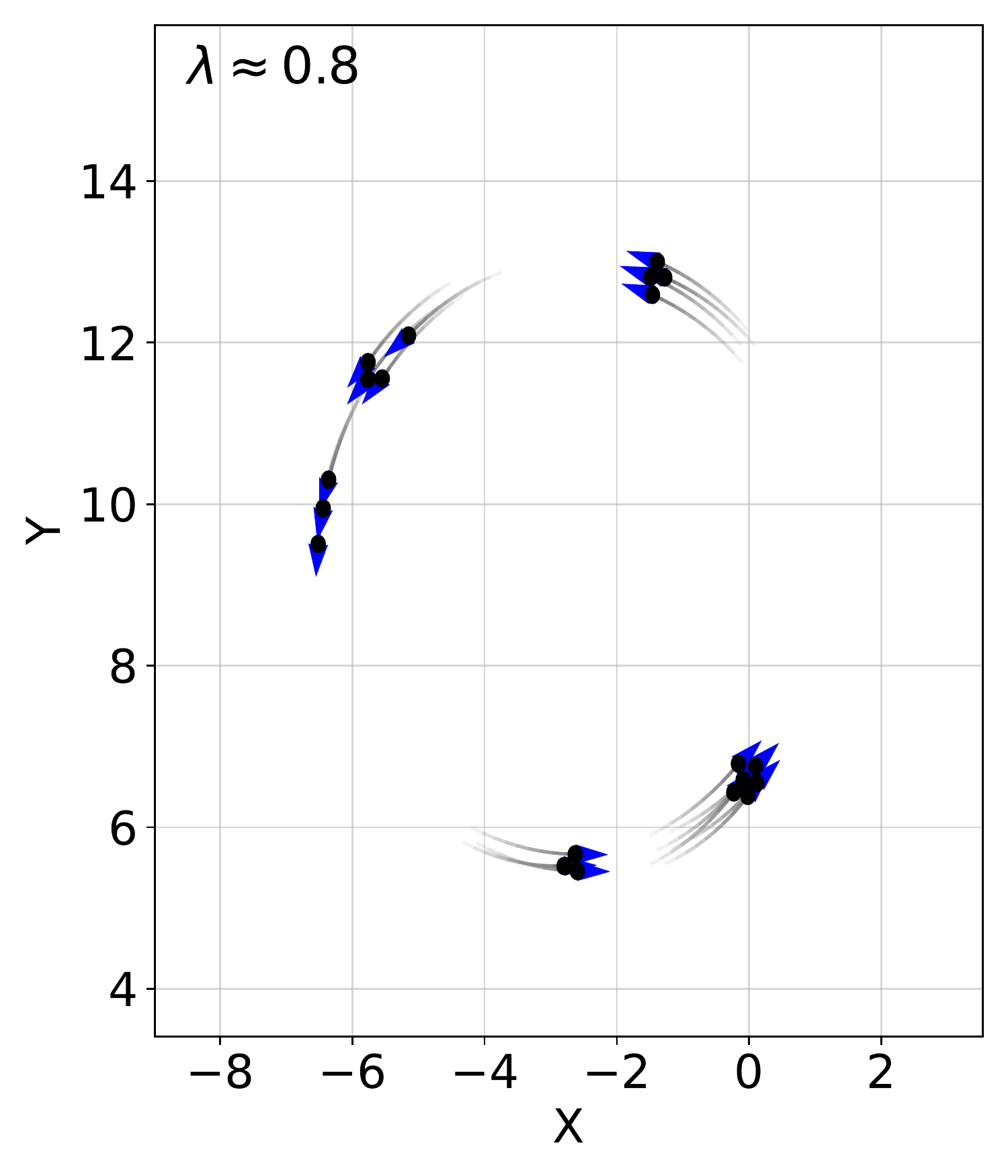}
	\end{subfigure}
	\begin{subfigure}[t]{0.16\textwidth}
		\includegraphics[width=\linewidth]{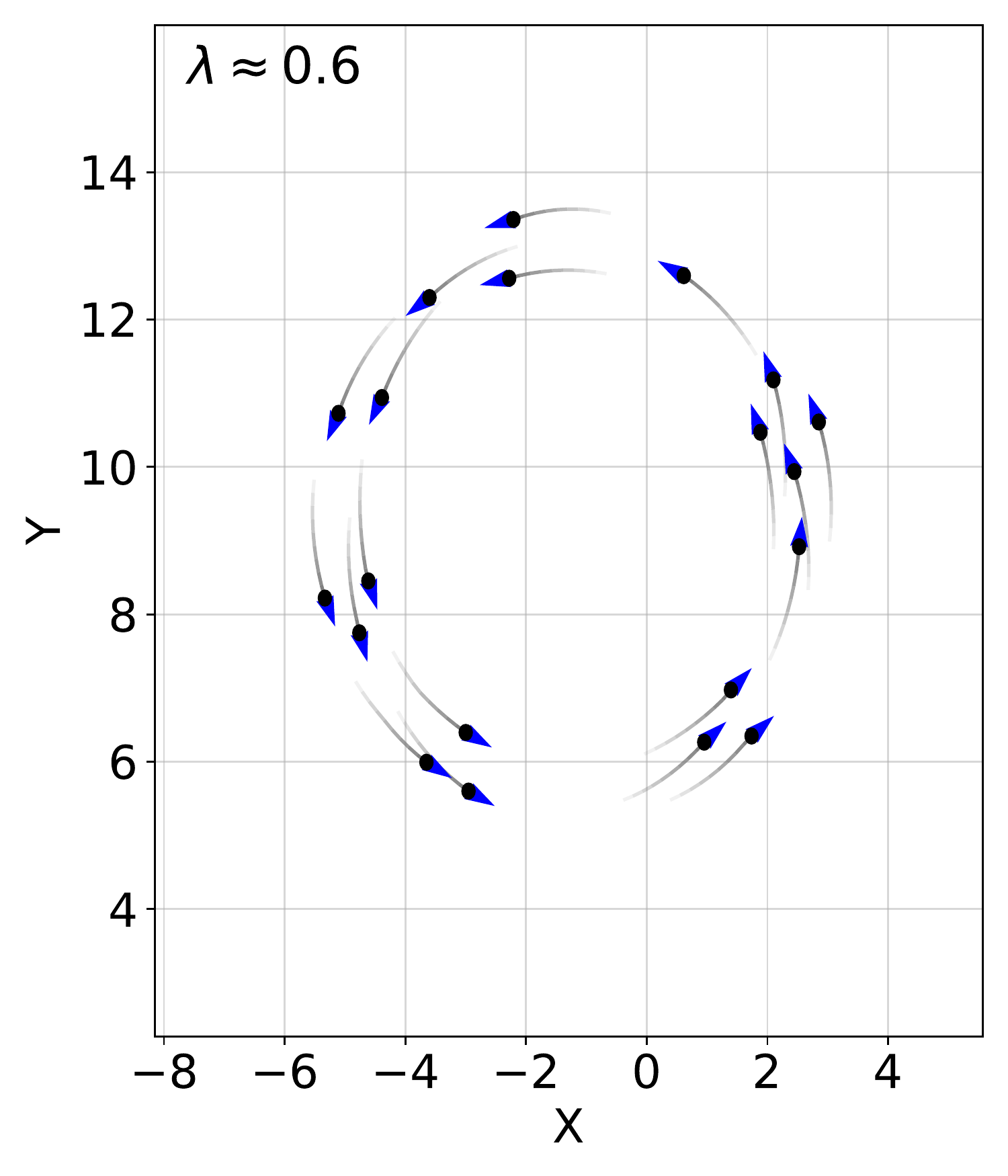}
	\end{subfigure}
	\begin{subfigure}[t]{0.16\textwidth}
		\includegraphics[width=\linewidth]{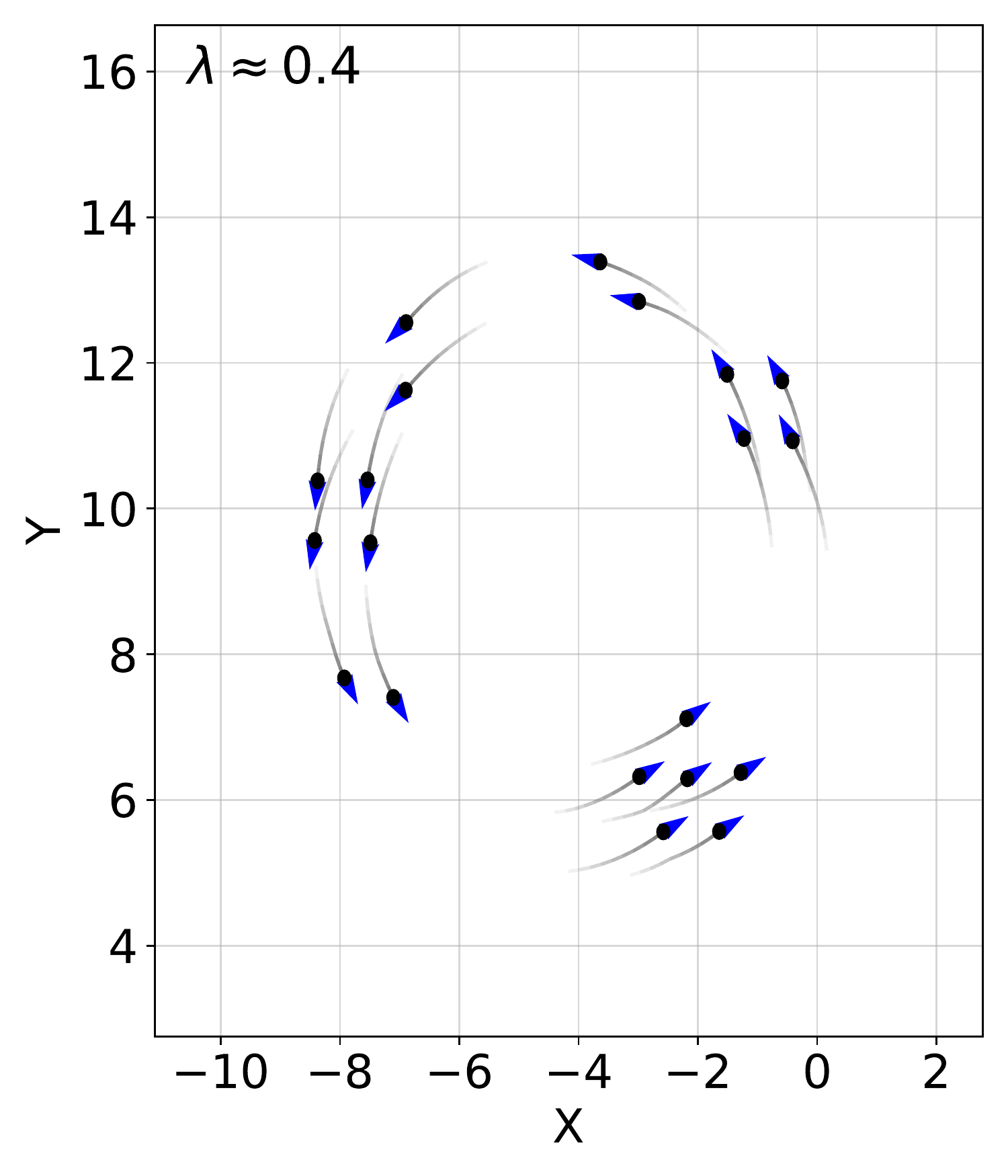}
	\end{subfigure}
	\begin{subfigure}[t]{0.16\textwidth}
		\includegraphics[width=\linewidth]{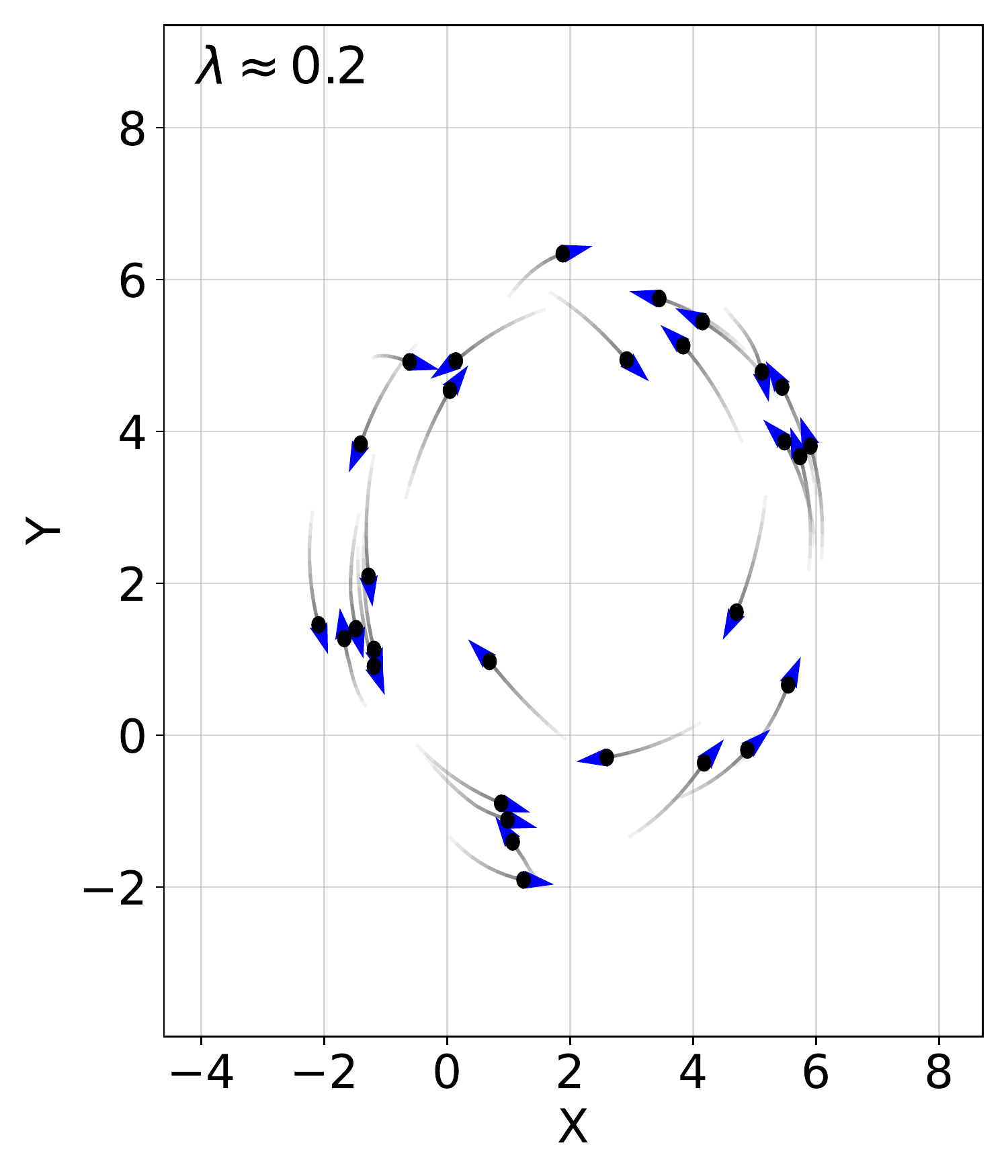} 
	\end{subfigure}
	\begin{subfigure}[t]{0.16\textwidth}
		\includegraphics[width=\linewidth]{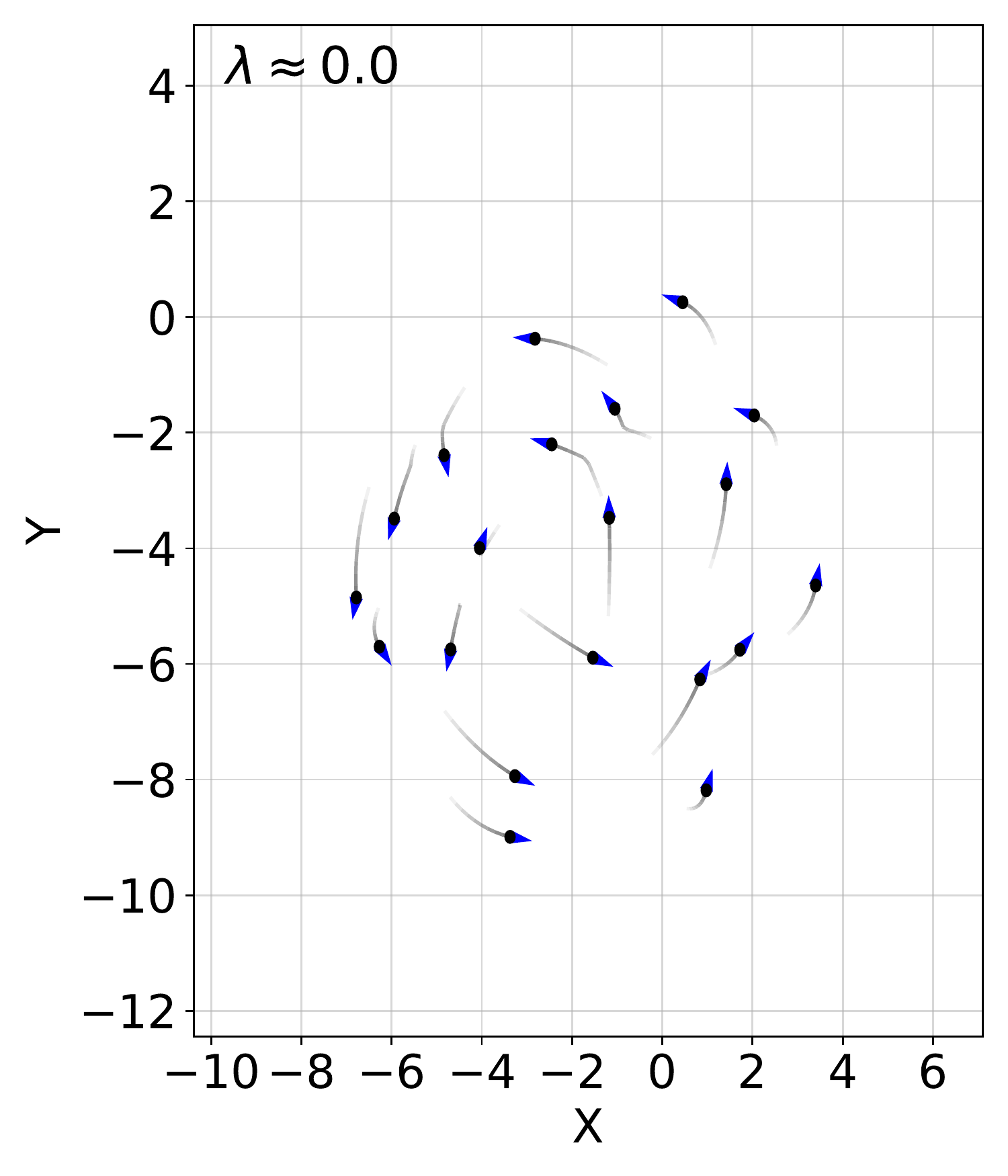}
	\end{subfigure}
\caption{\journaledit{Some examples of the ring quality metric $\convqual$. $\convqual = 1$ means all agents are evenly distributed around the same circle and moving tangent to it. As $\convqual \rightarrow 0$ we see the quality degrade: agents clump up and make the ring thicker or their velocities are not tangent to the circle. }}
\label{fig:lambdaexamples}
\end{figure*}

In \cref{sec:methodology:measurequality}, we describe our metrics to quantify the amount of interference caused by adding collision avoidance. 
\cref{sec:methodology:collisionavoidance} describes how we replicate each collision avoidance method and \cref{sec:methodology:simsetup} describes the particulars of the simulator setup. 

\subsection{Measuring Emergent Behavior Quality}
\label{sec:methodology:measurequality}

The metrics we introduce in our prior work \cite{taylor2020acc} consist of the ``fatness''  $\Fatness \in [0,1]$ and ``tangentness'' $\Tangentness \in [0,1]$, where the tangentness is similar to the normalized angular momentum in \cite{Chuang2007}. 
Formally, letting $\mu$ be the average position of all agents, i.e. $ \mu = \frac{1}{N} \sum_{i=1}^N \rb_i$, and $\rmin, \rmax$ be the minimum and maximum distance from any agent to the ring center, i.e. $\rmin = \min_{i \in \until{N}} \norm{\rb_{i} - \mu}$ and $\rmax = \max_{i \in \until{N}} \norm{\rb_{i} - \mu}$, 
the fatness and tangentness are defined (respectively) as
\begin{align}
	\Fatness(t) &= 
		1 - \frac{\rmin^2(t)}{\rmax^2(t)}, \quad
	\Tangentness(t) = 
		\frac{1}{N}
		\sum_{i=1}^{N} \bigg\lvert \frac{\rb_i - \mu}{\norm{\rb_i - \mu}} \cdot \frac{\dot{\rb}_i}{\norm{\dot{\rb}_i}} \bigg\rvert. 	
\end{align}
In other words, $\Fatness = 0$ implies a perfectly thin ring and $\Fatness = 1$ implies an entirely filled-in disc. 
$\Tangentness = 0$ represents perfect alignment between agents' velocities and their tangent lines and $\Tangentness = 1$ means all agents are misaligned. 
We also define
$\Fatmean(t), \Tangmean(t)$
as the average fatness, tangentness over the time interval $[t-T, t]$.
We then define a single metric $\convqual~\in~[0,1]$ as 
\begin{equation}
\convqual~=~1~-\max(\Fatmean, \Tangmean), 
\label{eqn:lambdadef}
\end{equation}
where $\convqual = 1$ represents a perfect ring and $\convqual = 0$ represents maximum disorder. 
\journaledit{\cref{fig:lambdaexamples} shows some examples.}

\subsection{Collision Avoidance}\label{sec:methodology:collisionavoidance}

\begin{table*}[]
\centering
\begin{tabular}{l|l|l}
\textbf{Strategy} &  \textbf{Meaning of $\cscale$}  & \textbf{Range of $\cscale$ tested}  \\ \hline
Potential \cite{Szwaykowska2016}  &  Repulsive force multiplier  & $[0,1]$  \\ \hline
Gyro \cite{DongEuiChang2003} & Repulsive force multiplier & $[0,1]$ \\ \hline
CBC \cite{Borrmann2015} & \makecell{Factor to slow down approach to barrier} & $[10^{-5}, 10^5]$ \\ \hline
ORCA \cite{VanDenBerg2011} & \makecell{Planning time horizon in seconds} & $[0.1, 10]$
\end{tabular}
\caption{A summary of each collision avoidance strategy and the range of cautiousness parameter $\cscale$ we test}
\label{table:eachstrategy}
\end{table*}

Here we summarize each collision avoidance technique that we replicate: potential fields \cite{Szwaykowska2016}, the ``gyro'' method \cite{DongEuiChang2003}, Optimal Reciprocal Collision Avoidance \cite{VanDenBerg2011}, and Control Barrier Certificates \cite{Borrmann2015}, which we refer to here simply as Potential, Gyro, ORCA, and CBC for brevity. 
In all cases, we replicate the decentralized version of each technique where agents are only aware of the positions and velocities of other agents in their local neighbor set $\Nset_i$ but not their inputs, i.e. agent $j$ cannot use agent $i$'s desired input $\udesire{i}$ from \cref{eqn:udesire} nor its actual input $\uactual{i}$ from \cref{eqn:uactual}, $j \neq i$. 
\par
All strategies each have their own scalar ``cautiousness'' parameter, which we have replaced with a universal tuning parameter $\cscale > 0$ where increasing $\cscale$ represents an increase in how aggressively agents try to avoid each other. 
\cref{table:eachstrategy} summarizes the meaning of $\cscale$ for each. 
ORCA and CBC have a safety distance parameter $D_s$ where each strategy guarantees $\norm{\pos{i} - \pos{j}} > D_s$ to satisfy constraint C1 from \cref{sec:probformulation:model}. 
We set $D_s = 2.1r$, the agent diameter plus a 5\% margin, to allow for numerical imprecision from discretization in our simulation. 

\subsubsection{Potential}
The first choice of collision avoidance is a potential-fields scheme presented with \cite{Szwaykowska2016}. It is based on the gradient of a potential function which we represent in our framework as
\begin{multline}
    \fnocrashargs = \udesire{i} +  \\
         \nabla_{\pos{i}}
         \sum_{(\pos{j},\vel{j}) \in \Nset_i} \cscale \exp\left(-2\frac{\norm{\pos{i} - \pos{j}}}{\lr}\right). 
\label{eqn:repulse}         
\end{multline}

\subsubsection{Gyro}
The second collision avoidance scheme we study is the ``gyroscopic'' force presented in \cite{DongEuiChang2003}. 
This produces a force orthogonal to the agents velocity that ``steers'' the agent without changing its speed. 
It can be written in closed-form as

\begin{equation}
\begin{gathered}
	\fnocrashargs = \udesire{i} + \\
	 R_{90\degree} \frac{\vel{i}}{\norm{\vel{i}}} \sgnz\left( (\pos{j}^* - \pos{i}) \times \vel{i} \right) \, U(\norm{\pos{i} - \pos{j}^*}), 
\end{gathered}
\label{eqn:gyroterm}
\end{equation}

where $R_{90\degree}$ is a $90\degree$ rotation matrix to give us a vector orthogonal to agent i's velocity, $(\pos{j}^*,\vel{j}^*) = \argmin_{(\pos{j},\vel{j}) \in \Nset_i} \norm{\pos{i} - \pos{j}}$ is the state of the \textit{nearest} agent
and 
\begin{equation}
	\sgnz(x) = \begin{cases}
		1 & x = 0 \\
		\sgn(x) & \text{otherwise }
	\end{cases}
\end{equation}
is the sign function modified such that the agent is forced to steer left during a perfect head-on collision, i.e. $\pos{j}^* - \pos{i}$ and $\vel{i}$ are pointing in the same direction. 
The cross product term in \cref{eqn:gyroterm} abuses notation somewhat; both vectors are 2D so we consider the result of the cross product here to be the scalar z component of the 3D cross product if both vectors were in the XY plane. 
The function $U(d)$ represents a potential controlling the magnitude of the steering force. 
As \cite{DongEuiChang2003} specifies, the magnitude $U(d)$ is arbitrary so we choose
\begin{equation}
	U(d) = 2\frac{\cscale}{\lr} \exp\left( -2 \frac{d}{\lr} \right)
	\label{eqn:forcemagnitude}
\end{equation}
such that the force magnitude is exactly the same as the method of potential fields \cref{eqn:repulse} with one other agent.

\subsubsection{Control Barrier Certificates (CBC)} \label{sec:cbc}

CBC \cite{Borrmann2015} uses a \textit{barrier function} $B_{ij}$ which is a function of the states $\pos{i}, \pos{j}, \vel{i}, \vel{j}$ of two agents and is defined such that $B_{ij} \rightarrow \infty$ as agent $i$ is about to collide with agent $j$. 
They define it as
\begin{equation}
	B_{ij} = 1 / \left( \frac{\rpos{ij}}{\norm{\rpos{ij}}} \cdot \rvel{ij} 
		+ \sqrt{4 \amax (\norm{\rpos{ij}} - D_s) }
	\right). 
\end{equation}
where $\rpos{ij} = \pos{i} - \pos{j}$ and $\rvel{ij} = \pos{i} - \pos{j}$ are the relative positions and relative velocities respectively for two agents $i,j$.

The controller tries to minimize the amount of correction each agent applies subject to the safety constraint
\begin{equation}
    \dot{B}_{ij} \leq \frac{1}{\cscale B_{ij}}, 
\label{eqn:cbc:constraint}
\end{equation}
where we substitute their tunable parameter with $\cscale$. 
Intuitively, \cref{eqn:cbc:constraint} reads: the rate at which two agents $i,j$ approach the ``barrier'' of a collision with each other must slow down the closer they are to the barrier. 
The cautiousness parameter $\cscale$ can scale their approach speeds. 
We can express CBC in our framework as
\begin{equation}
\begin{split}
    & \fnocrashargs = \\
    & \quad \argmin_{\uactual{i}} \norm{\uactual{i} - \udesire{i}}^2 \\
		& \quad \begin{split}
    			\text{subj to. } & \dot{B}_{ij} \leq \frac{1}{\cscale B_{ij}} \; \forall j \in \Nset_i, \\
       		                    & \norm{\uactual{i}}_{\infty} \leq \amax.  
        \end{split}
\end{split}
\label{eqn:cbc:fcdef}
\end{equation}
Since agents do not know each other's inputs, agent $i$ assumes $\udesire{j} = 0, j \neq i$, i.e. that agent $j$'s velocity is constant. 
\cref{eqn:cbc:fcdef} is a quadratic program which we solve using the operator splitting quadratic programming solver (OSQP) \cite{osqp,osqp-infeasibility}. 
If a solution does not exist, we make agent $i$ brake, i.e. $\uactual{i} = -\vel{i}$, as \cite{Wang2017a} recommends doing.  
\par 
To guarantee safety, CBC requires the sensing radius $\lr$ obeys
\begin{equation}
    \lr \geq D_s + \frac{1}{4 \amax} \left( \sqrt[3]{4 \cscale \amax} + 2 \vmax \right)^2,
\label{eqn:cbc:lr}
\end{equation}
where we choose $\vmax = 2 v_0$ from \cref{eqn:udesire} to allow flexibility in case neighboring agents violate their speed set-point $v_0$.

\subsubsection{Optimal Reciprocal Collision Avoidance (ORCA)}
\label{sec:orca}
This strategy is based on the \emph{velocity obstacle} \cite{Fiorini1998,VanDenBerg2011}, which is the set of all relative velocities that would result in a crash with another agent within $\cscale$ seconds in the future (we use our universal tuning parameter $\cscale$).
ORCA takes this one step further by introducing the \textit{reciprocal} velocity obstacle, that is, a more permissive velocity obstacle that assumes the other agent is also avoiding its velocity obstacle instead of just continuing in a straight line. 
Let $\ballo{\mathbf{x}}{\rho}$ represent the open ball centered at $\mathbf{x}$ of radius $\rho$, i.e.
\[
    \ballo{\mathbf{x}}{\rho} = \lbrace \mathbf{p} \in \reals^2 \; | \; \norm{\mathbf{p} - \mathbf{x}} < \rho \rbrace. 
\]
Let $\ORCA{i}{j}$ denote the set of safe velocities for agent $i$ assuming that agent $j$ is also using the ORCA algorithm. 
We then define
\begin{equation}
    \orcame{i}{\cscale} = \ballo{\mathbf{0}}{v_0} \cap \bigcap_{j \in \Nset_i} \ORCA{i}{j}
\end{equation}
which represents the set of safe velocities slower than the set-point speed $v_0$ for agent $i$ when considering \textit{all} of its local neighbors in $\Nset_i$. 
\par 
In our discussion of ORCA so far, we consider safe \emph{velocity} inputs, however our agent model in \cref{eqn:model} assumes \emph{acceleration} inputs. 
To get around this we: (1) convert each agent's desired acceleration into the new velocity that would result from it, i.e. $\vel{i} + \Delta t \udesire{i}$, (2) find the safe velocity $\vsafe$ using ORCA, then (3) find the input $\uactual{i}$ necessary to achieve this new velocity. 

In summary, the definition of ORCA expressed in our framework is

\begin{equation}
\begin{split}
    \fnocrashargs = \frac{\vsafe - \vel{i}}{\Delta t} \\ 
    \vsafe = \argmin_{\mathbf{v} \in \orcame{i}{\cscale}} 
        \norm{\mathbf{v} - (\vel{i} + \Delta t \udesire{i})}
\end{split}
\label{eqn:orcdef}
\end{equation}
where $\Delta t$ is the timestep of the Euler integration in our simulator. 
Provided $\Delta t$ is sufficiently small, we find varying it does not change the results.

\subsection{Simulator Setup}
\label{sec:methodology:simsetup}

To integrate the dynamics equations of \cref{eqn:uactual}, we use Euler integration with the timestep $\Delta t$ set as 
\[
	\Delta t = \min\left(\frac{r}{\vmax}, 0.015\right)
\]
which ensures that no agent moves more than half its body length in order for the collision detection to function properly. We assume $\vmax = 2 v_0$ here which give agents some flexibility to move up to twice their set-point speed $v_0$. 
Capping $\Delta t$ at 0.015 ensures accuracy in the cases where $r$ is large. 
\par 
We initialize all agents on a spiral grid formation, where each agent is spaced at a distance of $\max( 5 r, \lr )$ to ensure agents do not collide at the start. 
We initialize their velocities such that $\norm{\vel{i}} = v_0$ with their initial headings chosen randomly. 

\par 
In all our experiments, we let the swarm stabilize for 10,000 simulated seconds then record metrics for an additional 2,000 seconds, i.e. the terminal time is $t=12000$ and the interval over which we measure the average ring quality $\convqual$ is $T=2000$ from \cref{sec:methodology:measurequality}. Potential fields and Gyro take longer to stabilize, so we use $t=32000, T=2000$ for those. 
\par

\section{Results} \label{sec:results}

\begin{figure*}
	\begin{subfigure}[t]{0.16\textwidth}
		\includegraphics[width=\linewidth]{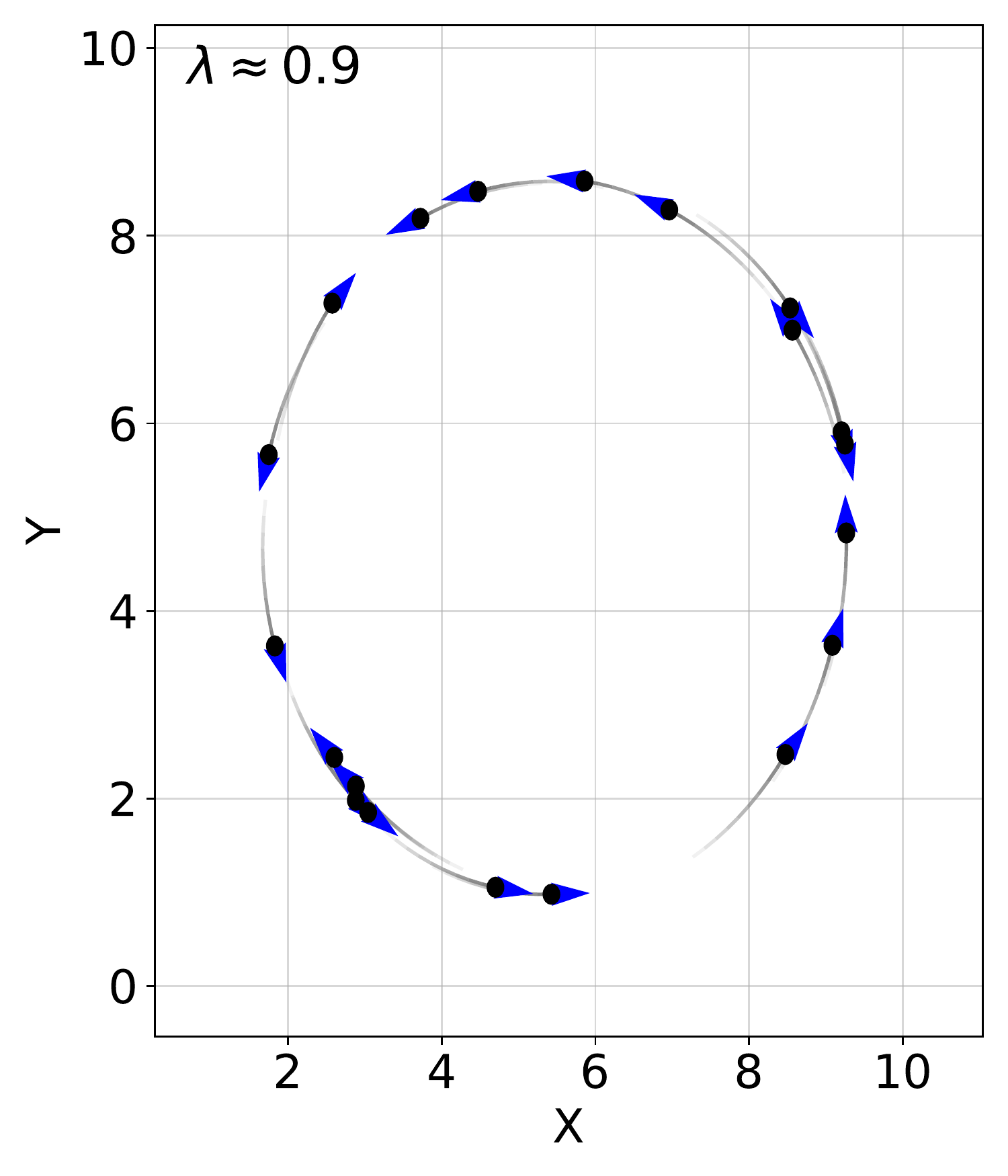}
		\caption{Perfect ring with collisions disabled}
		\label{fig:snapshots:perfect}
	\end{subfigure}\hfill
	\begin{subfigure}[t]{0.16\textwidth}
		\includegraphics[width=\linewidth]{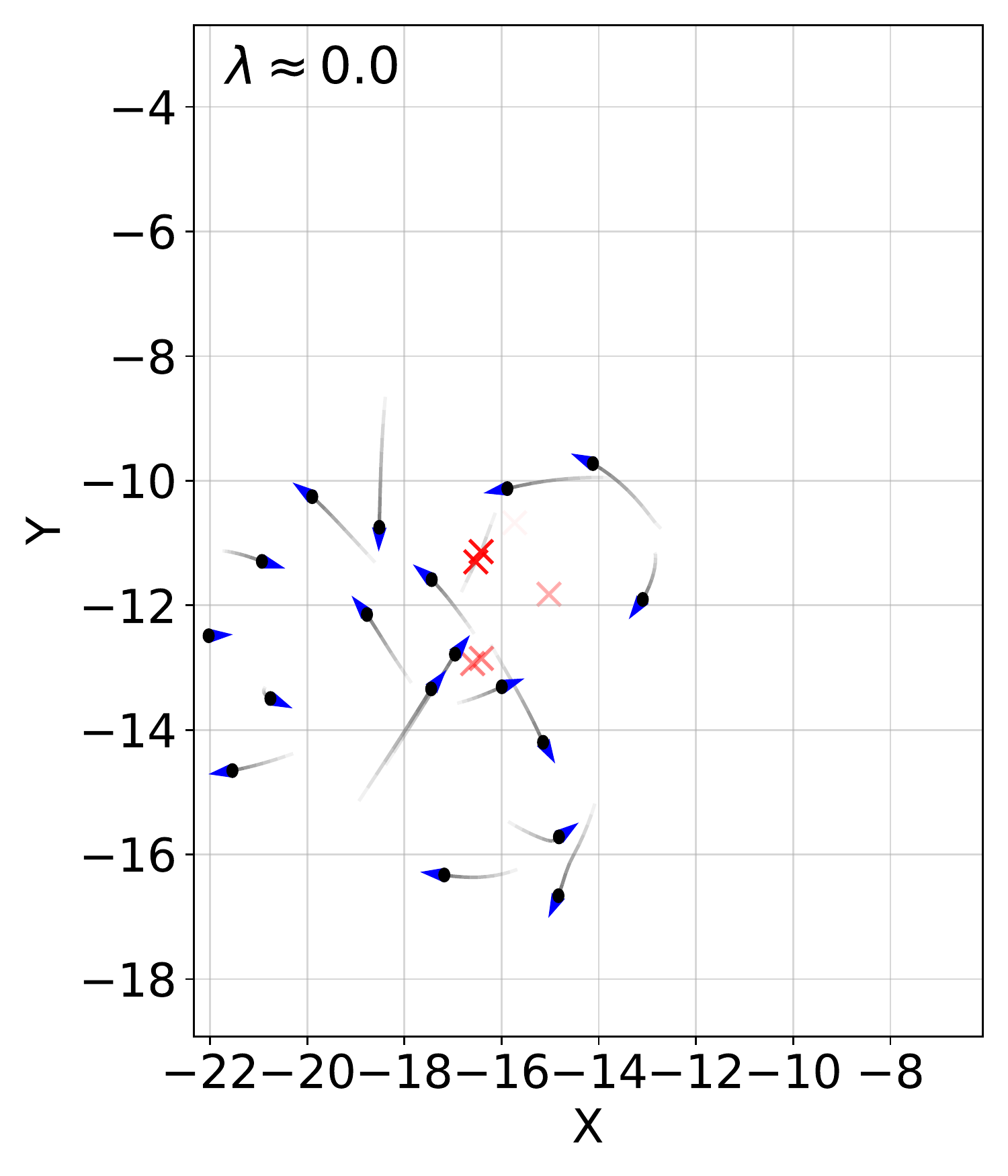}
		\caption{Collisions from insufficient collision avoidance}
		\label{fig:snapshots:collisions}
	\end{subfigure}\hfill
	\begin{subfigure}[t]{0.16\textwidth}
		\includegraphics[width=\linewidth]{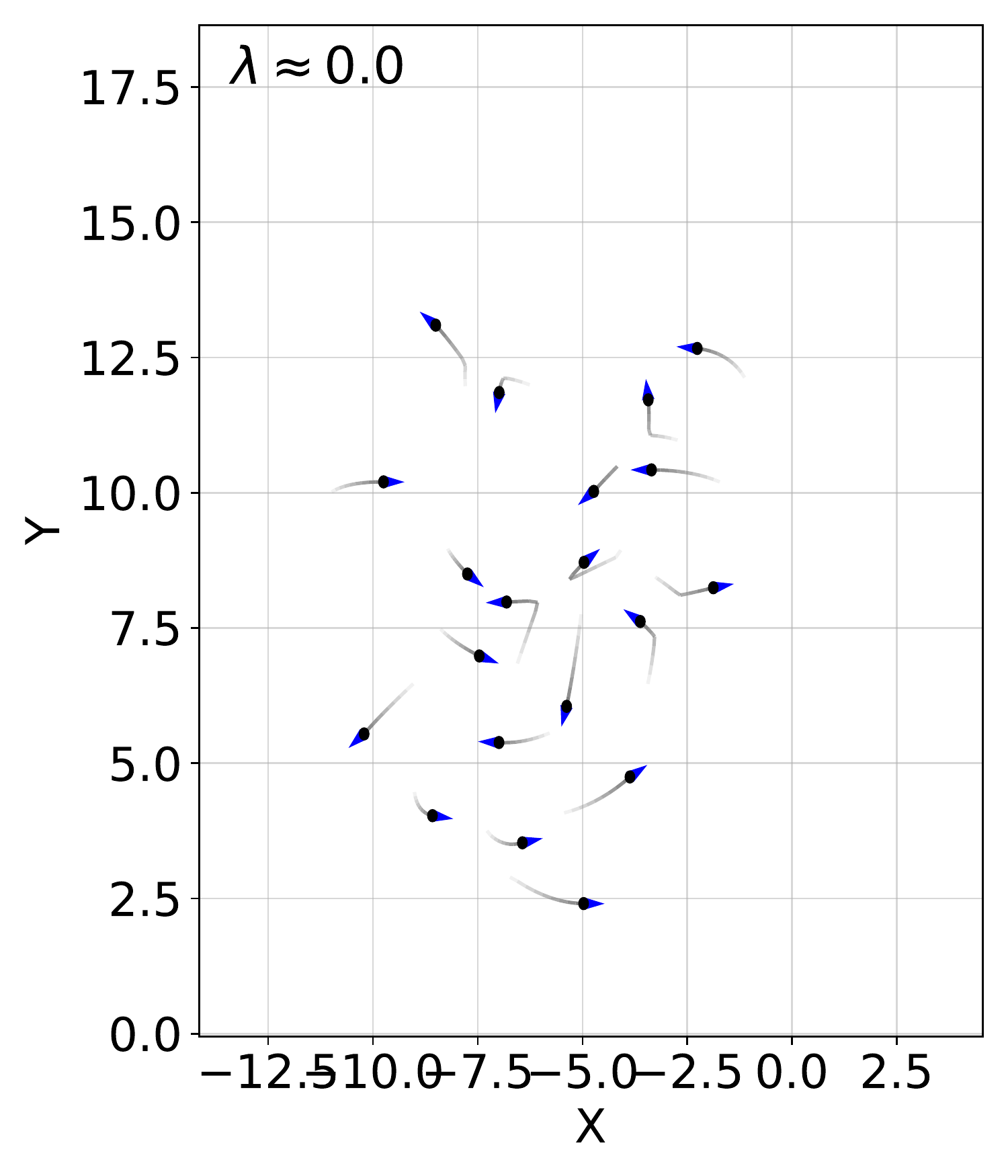}
		\caption{Excessive collision avoidance}
		\label{fig:snapshots:potentialaimless}
	\end{subfigure}\hfill
	\begin{subfigure}[t]{0.16\textwidth}
		\includegraphics[width=\linewidth]{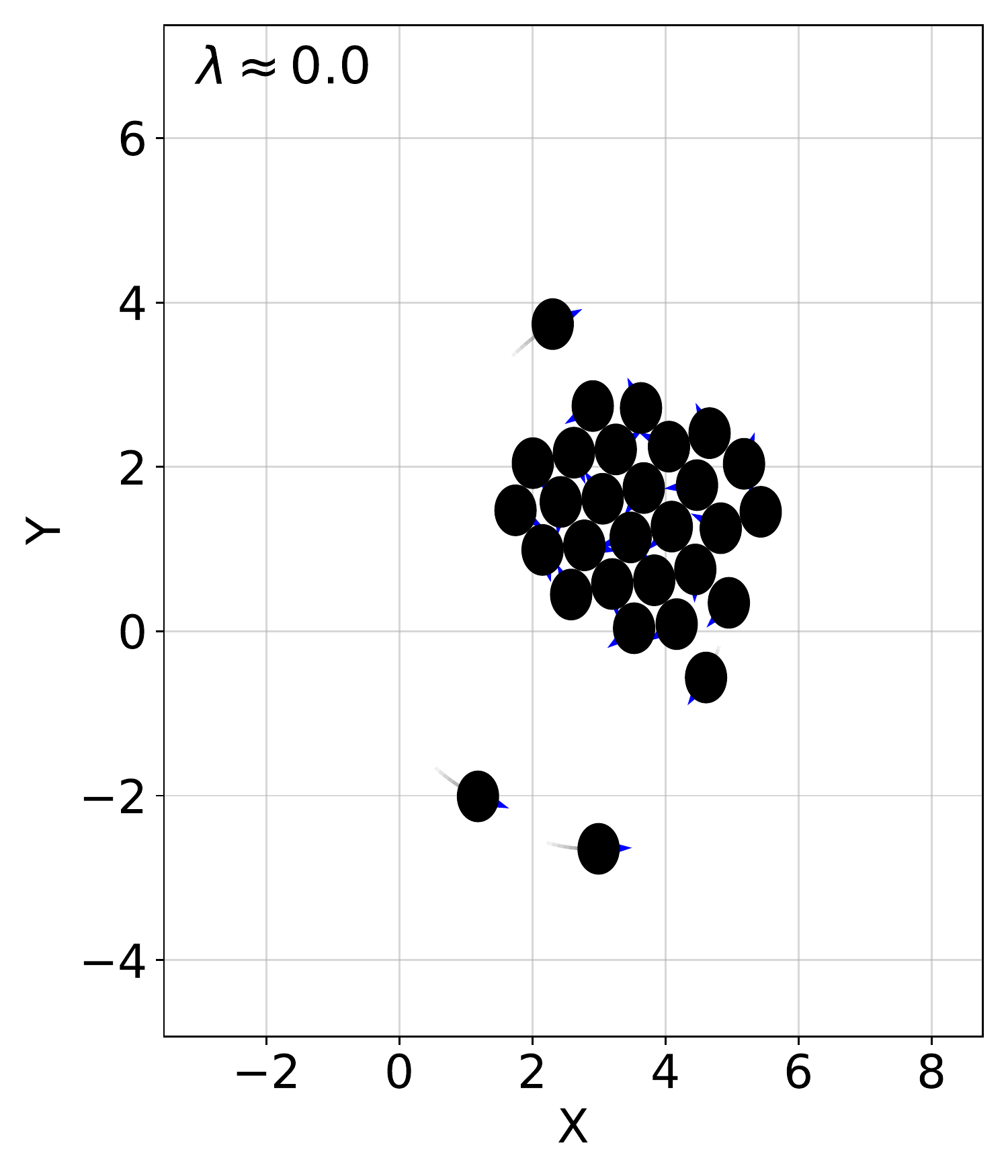}
		\caption{A deadlock with CBC}
		\label{fig:snapshots:cbcdeadlock}
	\end{subfigure}\hfill
	\begin{subfigure}[t]{0.16\textwidth}
		\includegraphics[width=\linewidth]{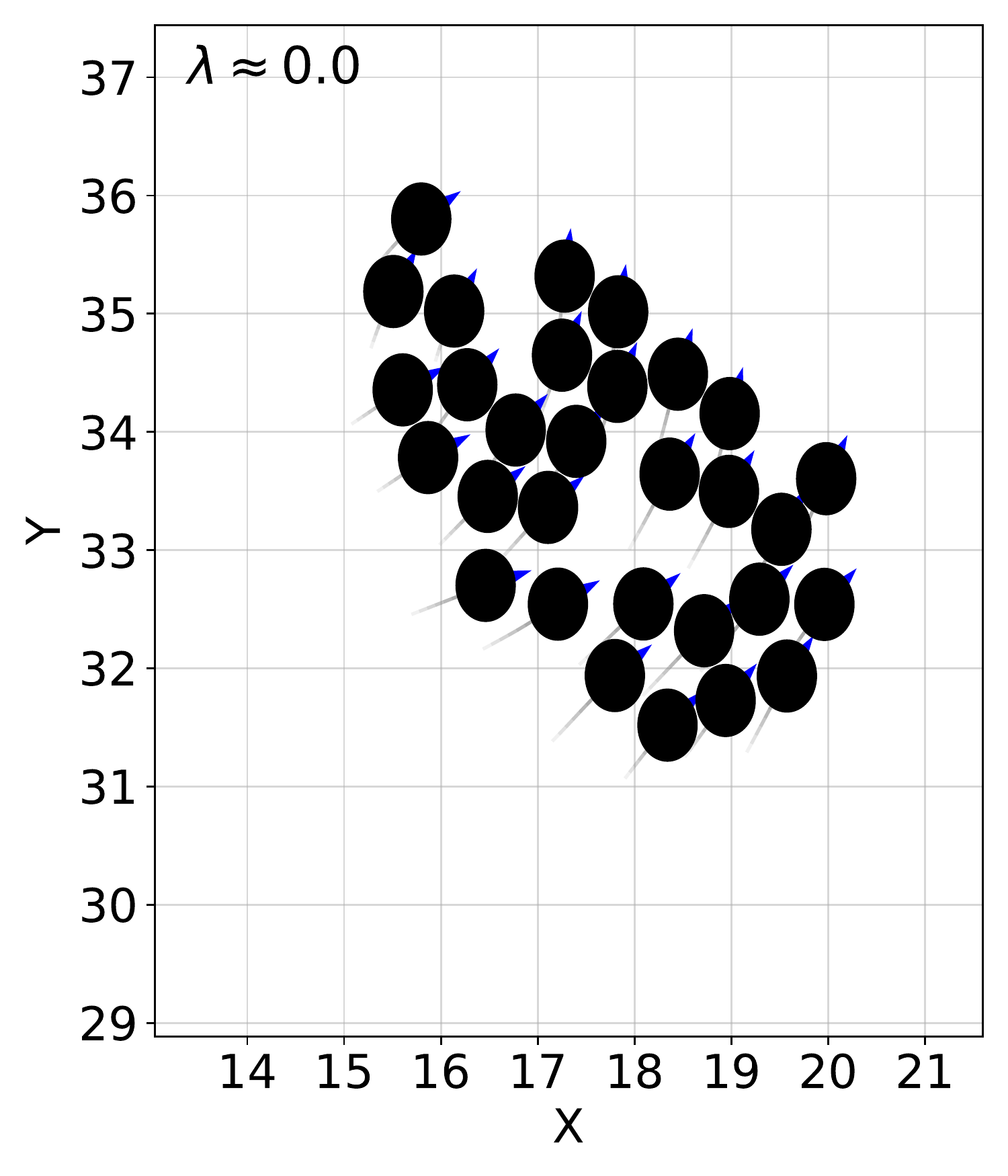}
		\caption{ORCA causing an ``unplanned flock''}
		\label{fig:snapshots:orcaflock}
	\end{subfigure}\hfill
	\begin{subfigure}[t]{0.16\textwidth}
		\includegraphics[width=\linewidth]{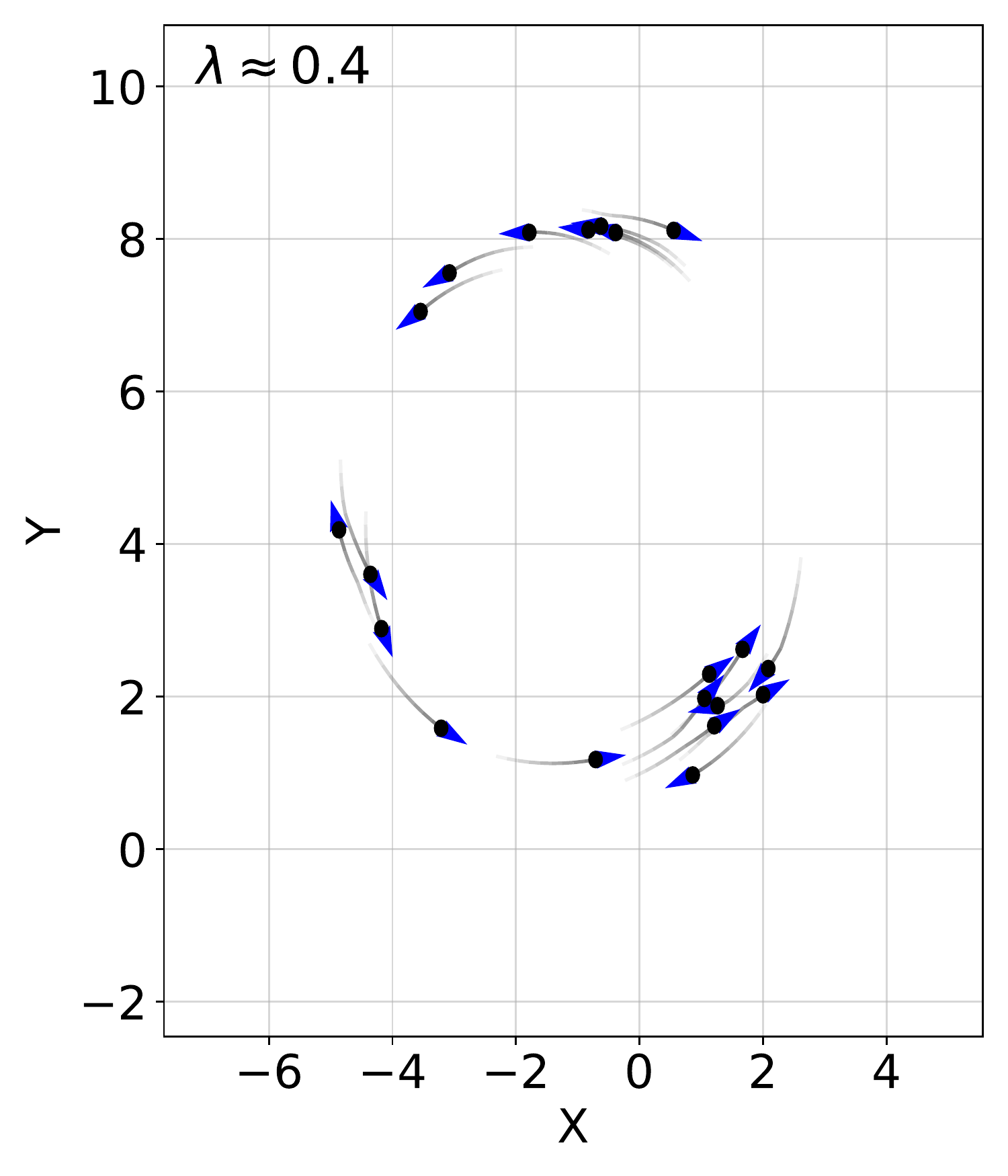}
		\caption{Attempted double-milling with ORCA}
		\label{fig:snapshots:orcadoublemill}
	\end{subfigure}
\caption{Different outcomes for the ring behavior depending on collision avoidance and parameters chosen. We include video versions of these as Supplemental Material. }
\label{fig:snapshots}
\end{figure*}

\begin{figure*}
	\begin{subfigure}[t]{0.25\textwidth}
		\includegraphics[width=\linewidth]{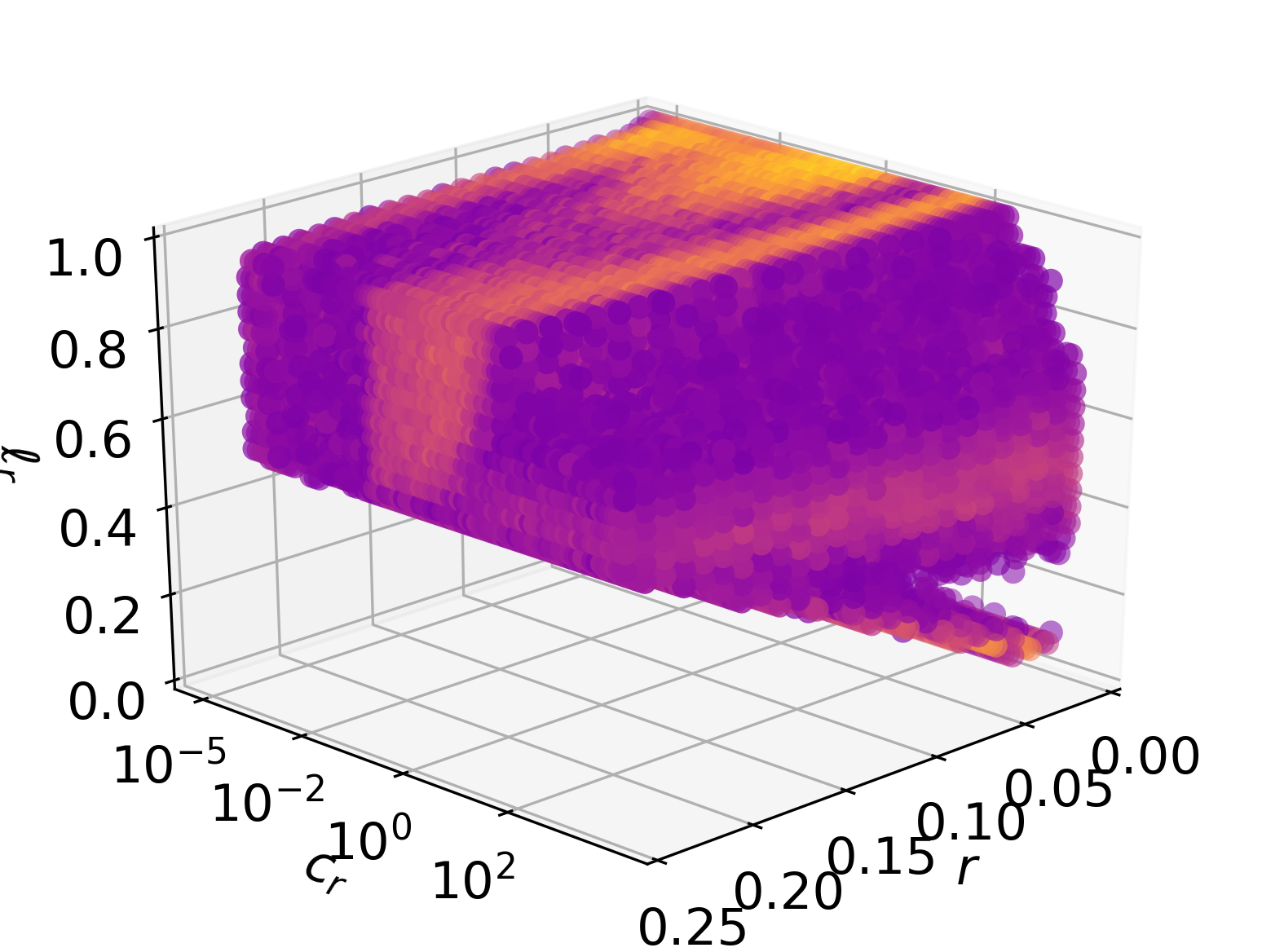}
		\caption{CBC}
		\label{fig:threed:cbc}
	\end{subfigure}\hfill
	\begin{subfigure}[t]{0.25\textwidth}
		\includegraphics[width=\linewidth]{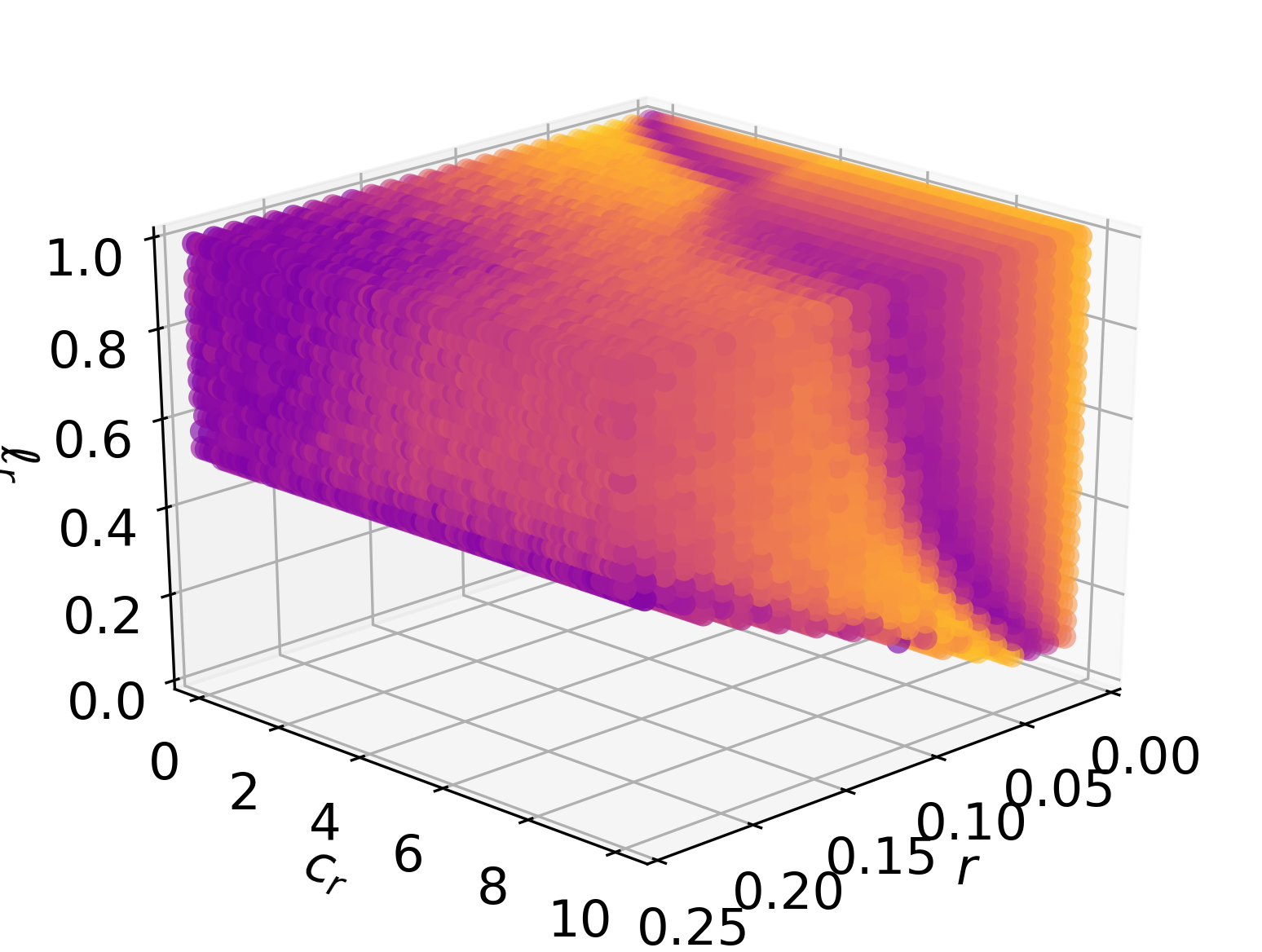}
		\caption{ORCA}
		\label{fig:threed:orca}
	\end{subfigure}\hfill
	\begin{subfigure}[t]{0.25\textwidth}
		\includegraphics[width=\linewidth]{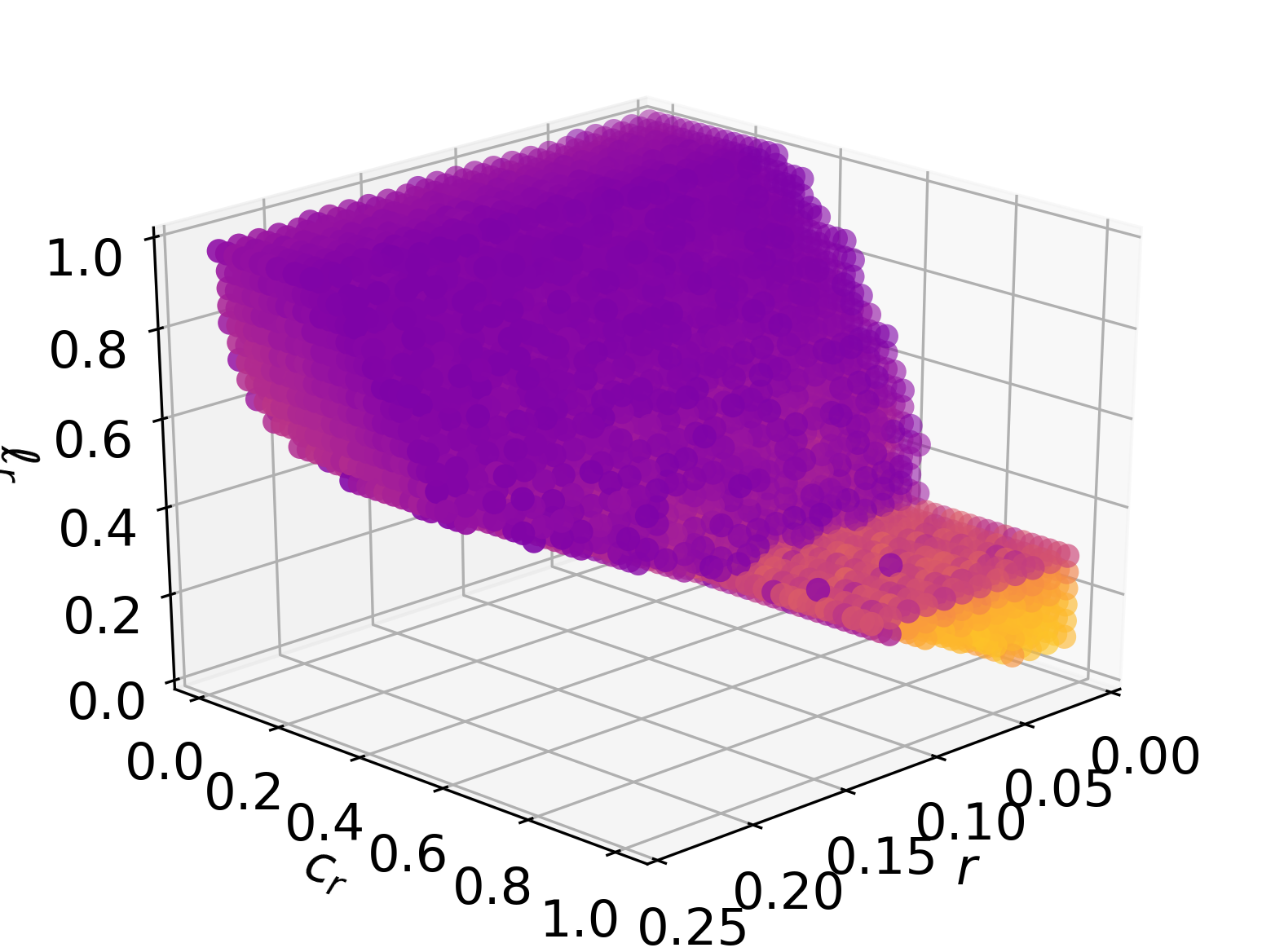}
		\caption{Potential}
		\label{fig:threed:potential}
	\end{subfigure}\hfill
	\begin{subfigure}[t]{0.25\textwidth}
		\includegraphics[width=\linewidth]{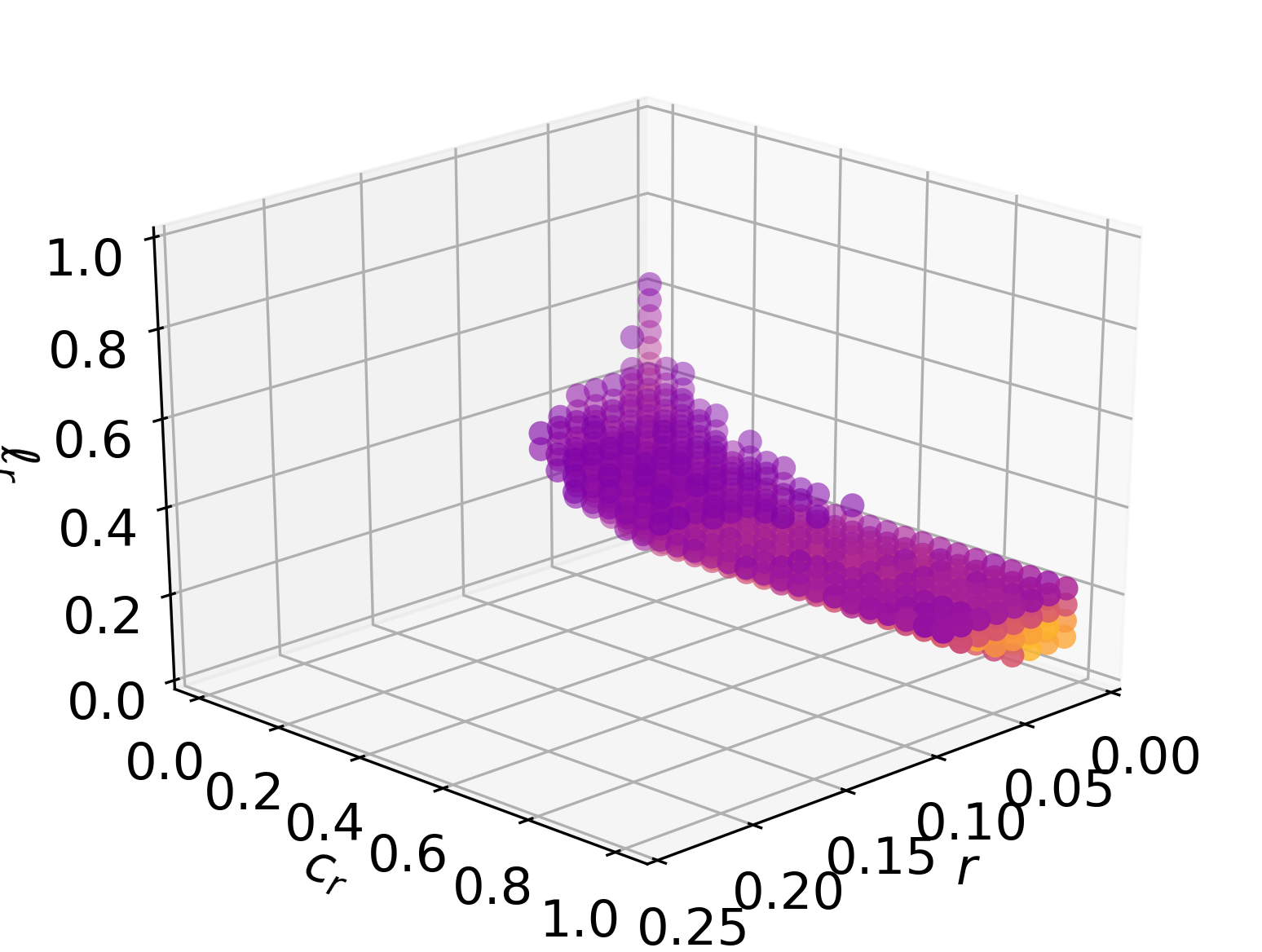}
		\caption{Gyro}
		\label{fig:threed:gyro}
	\end{subfigure}\hfill
\caption{Ring quality $\lambda$ as a function of $r, \cscale, \lr$ across all four collision avoidance algorithms. For ease of viewing we omit data points where $\convqual < 0.25$. }
\label{fig:threed}
\end{figure*}

\begin{figure}
	\includegraphics[width=\linewidth]{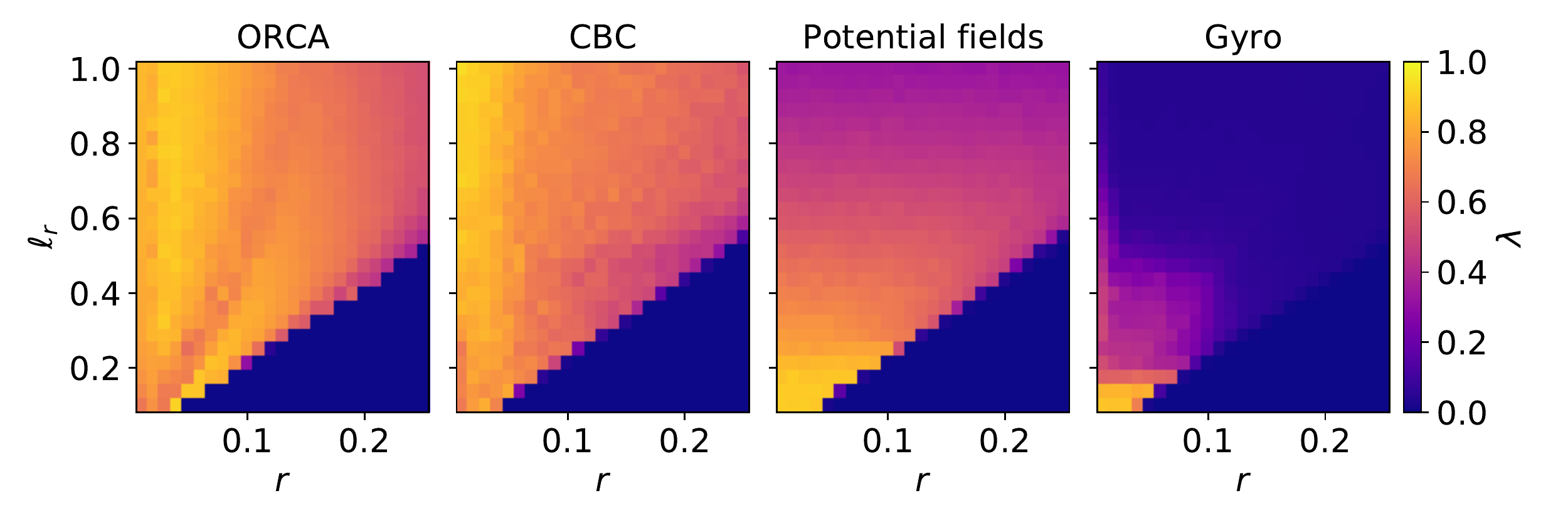}
	\caption{`Flattened' versions of \cref{fig:threed} where we choose the best value of $\convqual$ 
	along the $\cscale$ dimension}
	\label{fig:threedflat}	
\end{figure}

\begin{figure}
	\centering
	\includegraphics[width=0.8\linewidth]{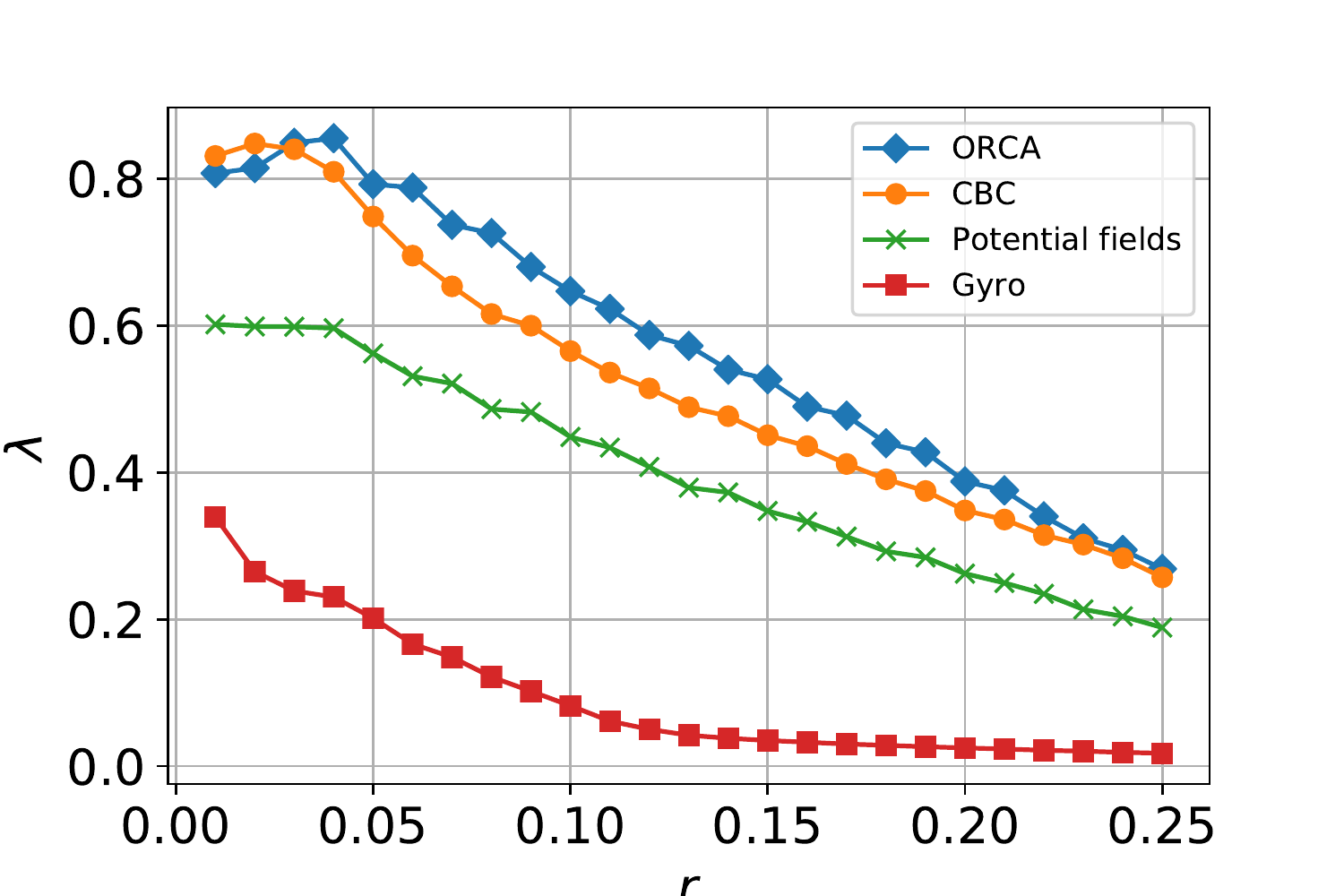}
	\caption{The ring quality $\convqual$ as a function of the agent size, where $\convqual$ is averaged over the sensing radius $\lr$. \journaledit{A further `flattened' version of \cref{fig:threedflat}}}
	\label{fig:r:lmb:lines}		
\end{figure}

In this section, we use our ring quality metric $\convqual$ (\cref{eqn:lambdadef}) to assess the interference caused by each collision avoidance algorithm. 
\cref{sec:results:interference} gives an overview of the interference caused by each algorithm under a fixed number of agents and a large sampling of the parameter space, where we ``tune'' each collision avoidance algorithm through its respective cautiousness parameter $\cscale$ to fairly represent each one at its best-case performance. 
\cref{sec:results:tuningdifficulty} explores the tuning process in more detail by showing examples of specific choices of $\cscale$. 
Finally, \cref{sec:results:orcacbc:scalability} explores varying the number of agents to check how the interference scales. 

\subsection{Collision Avoidance Interference}
\label{sec:results:interference}

\cref{fig:snapshots} shows a sampling of the interference caused by each collision avoidance algorithm.
\cref{fig:snapshots:perfect} shows a perfect ring with collisions and collision avoidance disabled.   
If we add the constraints and collision avoidance, the interference manifests in unique and unpredictable ways. 
Potential or Gyro, neither of which carry safety guarantees, may allow agents to crash and respawn if $\cscale$ is too low (\cref{fig:snapshots:collisions}), or scatter aimlessly if $\cscale$ is too high (\cref{fig:snapshots:potentialaimless}). 
Also noteworthy is when the agent size $r$ is too large: CBC will simply cause most agents to deadlock in a large clump (\cref{fig:snapshots:cbcdeadlock}) whereas ORCA keeps agents moving but in an ``unplanned flock'' instead of a ring (\cref{fig:snapshots:orcaflock}). 
\par 
We note from our previous work \cite{taylor2020acc} that each collision avoidance algorithm needs to be tuned to function optimally. 
In particular, the cautiousness parameter $\cscale$ (\cref{sec:methodology:collisionavoidance}) has different meanings and units across all four algorithms, so we make an effort to explore many choices of $\cscale$ for the most fair comparison. 
Since we are mainly interested in the effects of agent size and collision avoidance on the swarm behavior, we vary two other parameters in addition to $\cscale$ which we consider to be the most critical for avoiding collisions. 
First, we consider the agent size $r$ to understand the effect that physical size has on the swarming behavior. 
Second, we consider the sensing radius $\lr$, since this is arguably one of the most critical parameters to guarantee safety since it directly controls how much information agents have about imminent collisions. 
We choose 25 unique parameters for both $r, \lr$ such that they evenly span the ranges $r \in [0, 0.25], \lr \in [0, 1]$.
To tune each collision avoidance, we choose 100 values for $\cscale$ in ranges shown in \cref{table:eachstrategy}. 
We feel these ranges exhaust the possibilities for each algorithm. 
For each $r,\cscale,\lr$ triplet, we use 50 seed values representing different initial conditions and take an average across all the seeds. 
We hold the other parameters fixed at $N = 20.0, \alpha = 0.001, \timedelay = 2.5, \beta = 1.0, v_0 = 0.12, \amax = 0.6$. 
\par
\cref{fig:threed} shows the results for all four algorithms as 3D point clouds, where we only show points where $\lambda > 0.25$. 
\cref{fig:threedflat} shows a `flattened' version of \cref{fig:threed} where we select the value of $\cscale$ that produces the highest $\convqual$. In other words, \cref{fig:threedflat} shows the best possible performance of each algorithm under agent size $r$ and sensing radius $\lr$ constraints. 
\par 
\cref{fig:r:lmb:lines} further ``flattens'' \cref{fig:threedflat} by averaging across the $\lr$ dimension. 
It is clear that despite tuning each algorithm to the best of its abilities, the effect of the agent size on performance is unavoidable.

\subsection{Tuning Difficulty}
\label{sec:results:tuningdifficulty}

\begin{figure}
	\includegraphics[width=\linewidth]{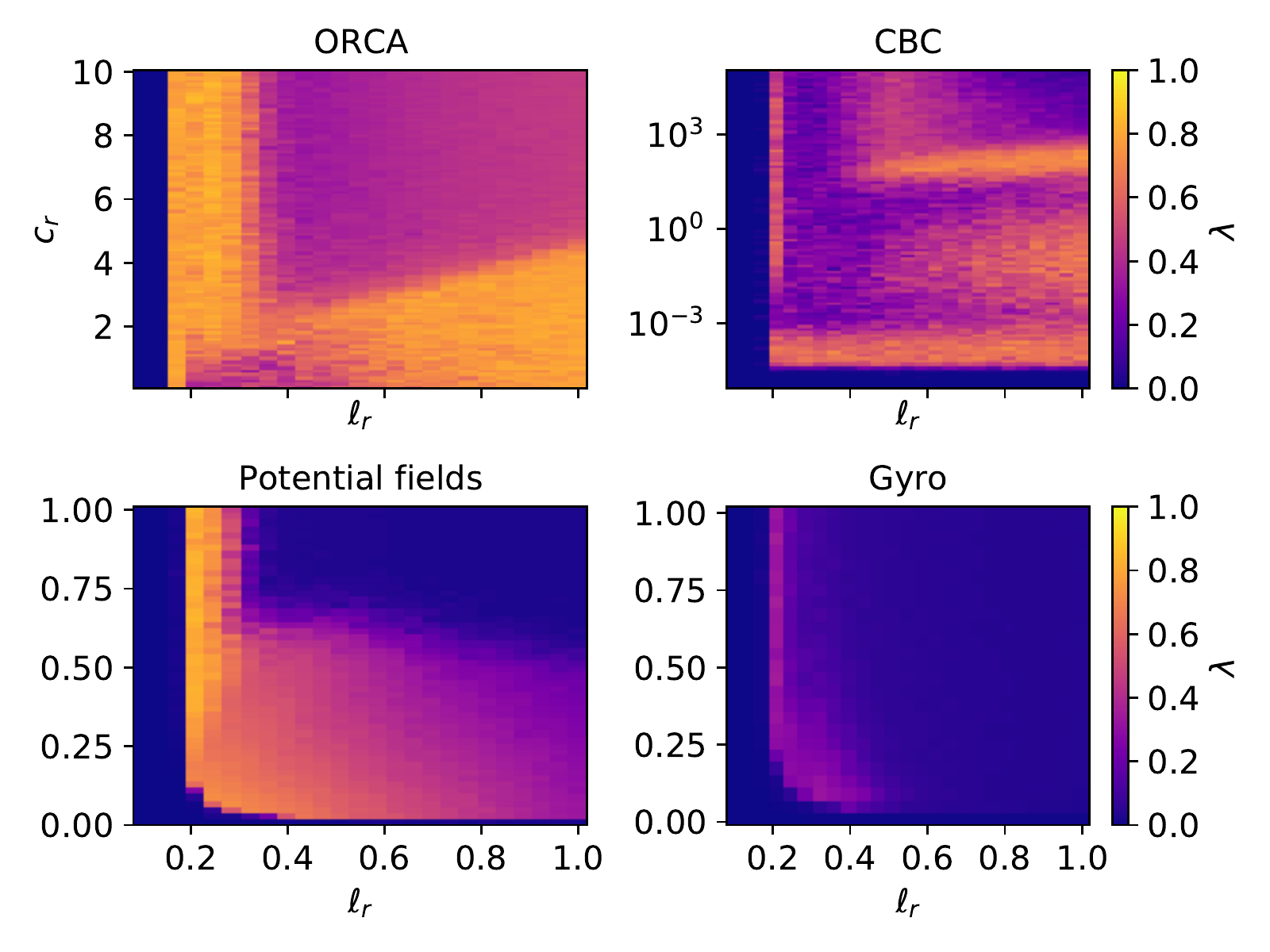}
	\caption{How the cautiousness parameter $\cscale$ and sensing radius $\lr$ affect $\convqual$ for $r = 0.8$}
	\label{fig:tuningdifficulty}
\end{figure}

In \cref{sec:results:interference} we eliminate the variable $\cscale$ choosing the value of $\cscale$ that results in the highest ring quality $\convqual$. 
We accomplish this by exhaustively searching the space of $\cscale$ over ranges appropriate to each algorithm, but it is worth examining what this space looks like. 
\cref{fig:tuningdifficulty} shows an example of slices of \cref{fig:threed} by fixing the agent size $r = 0.8$. 
One could read this as follows: if $r, \lr$ are fixed, then \cref{fig:tuningdifficulty} shows a ``map'' of how to select the best $\cscale$ that results in the best ring quality. 
However, owing to the multiple local minima with ORCA or CBC (especially CBC), tuning these as part of a larger deployment would require a global optimization strategy with large amounts of computing power. 
Indeed, our results seem to concur with \cite{Vasarhelyi2018a}: to tune the parameters of their swarming system (where some of the parameters control collision avoidance) they require days of supercomputer time using an evolutionary algorithm.

\subsection{Scalability}
\label{sec:results:orcacbc:scalability}

\begin{figure}
	\includegraphics[width=\linewidth]{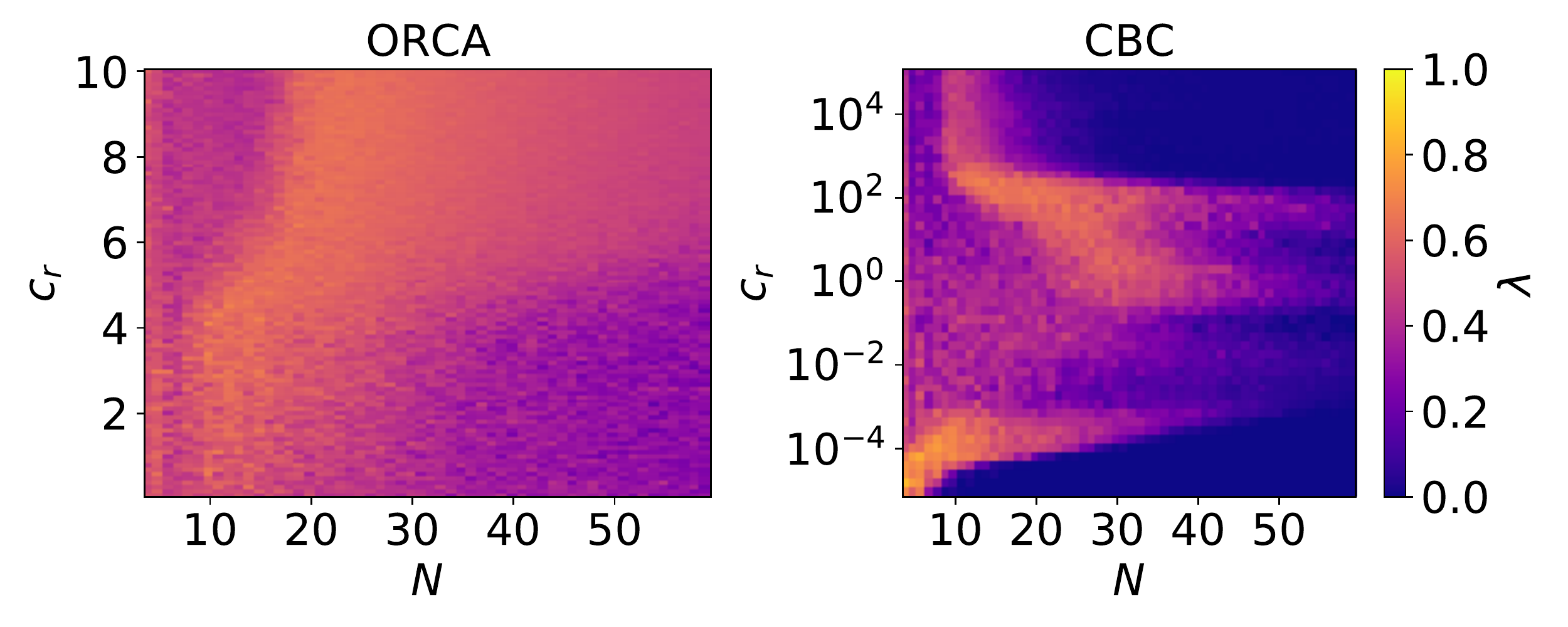}
	\caption{The effect of the number of agents $N$ and cautiousness parameter $\cscale$ on the ring quality.}
	\label{fig:compare:agentcount}
\end{figure}

We turn our attention to just ORCA and CBC, since as \cref{fig:r:lmb:lines} shows, their performance seems consistently better than Gyro or Potential. 
Since our previous results only use $N=20$, we now determine how the interference manifests with different numbers of agents. 
\par 
\cref{fig:compare:agentcount} shows a comparison of the quality $\lambda$ of ORCA and CBC as we vary the number of agents, again using the same range of the cautiousness $\cscale$ described in \cref{table:eachstrategy}. 
For each selection of parameters we run 50 trials with random initial velocities. 
We fix the other parameters at $\alpha = 0.001, \timedelay = 2.5, \beta = 1.0, v_0 = 0.12, \ell_r = 1.0, \amax = 0.6, r = 0.15$. 
Similar to \cref{fig:tuningdifficulty}, the parameter space is quite nonintuitive. For many selections of $\cscale$, the performance \textit{increases} with $N$ then starts decreasing again. 
This seems similar to the ``increase/decrease'' scenario described in \cite{hamann2020guerrilla}, where swarm agents at first cooperate with each other for low $N$ but begin to interfere once $N$ increases. 
It is only by adding constraints of collisions and collision avoidance that we notice such an effect. 
Otherwise, there is no mechanism for agents to interfere with each other and the performance is infinitely scalable.

\section{Conclusion}

Clearly, we cannot expect swarming algorithms in general to behave the same once we add constraints of agent size and damaging collisions, as this illustrative example shows. 
In order to maximize the swarm's performance under such constraints, there seems to be a difficult and nonintuitive tuning process involving many nonlinearities and local-maxima. 
We recommend that swarming behaviors which are to be deployed on platforms exhibiting the constraints shown here should be co-designed with collision avoidance, as opposed to designing them independently then combining them which might incur a lengthy tuning process and unexpected side effects. 
Our results suggest that future research is needed to understand how to effectively deploy existing swarming behaviors under collision constraints. 
\par 
Such future research might take a number of directions.
It might be worth studying the parameter tuning process in more depth since it is an enormously computationally expensive part of any swarm algorithm deployment, for instance as \cite{carolus2020control} does to find the most important parameters on which to focus tuning effort. 
It is also possible, as \cref{fig:r:lmb:lines} suggests, that there is an upper limit to how well a swarming behavior can function given limits on agents' sensing, their physical size, and maneuvering abilities.
It is possible that in future work we can co-design swarming algorithms to achieve this upper limit dynamically without exhaustive parameter tuning.

\section*{\journaledit{Acknowledgments}}
This work was supported by the Department of the Navy, Office of Naval Research (ONR), under federal grants N00014-19-1-2121 and
N00014-20-1-2042.
The experiments were run on ARGO, a research computing cluster provided by the Office of Research Computing at George Mason University, VA. (\url{http://orc.gmu.edu}).

\bibliographystyle{elsarticle-num}

 \end{document}